%% file: bmvc_review.tex
\pdfoutput=1
\documentclass{bmvc2k}

\input{preamble}
\usepackage{colortbl}
\usepackage{arydshln}
\usepackage{multirow}
\usepackage{tabu}
\usepackage{makecell}
\usepackage{threeparttable}
\usepackage{graphicx}
\usepackage{wrapfig}

\usepackage{booktabs}   
\usepackage{siunitx}    
\usepackage{etoolbox}   
\usepackage[most]{tcolorbox}

\usepackage{textcomp}

\definecolor{BoxBackground}{RGB}{240, 240, 240} 
\definecolor{BoxFrame}{RGB}{0, 0, 0} 
\definecolor{TitleBackground}{RGB}{0, 0, 0} 
\definecolor{TitleText}{RGB}{255, 255, 255} %

\tcbset{
  academicbox/.style={
    boxsep=5pt,
    left=2pt,
    right=2pt,
    bottom=0.5pt,
    boxrule=0.5pt,
    colback=BoxBackground,
    colframe=BoxFrame,
    colbacktitle=TitleBackground,
    coltitle=TitleText,
    enhanced,
    attach boxed title to top left={yshift=-0.1in,xshift=0.1in},
    boxed title style={boxrule=0pt,colframe=white},
    title={#1},
  }
}
\newtcolorbox{AcademicBox}[1][]{academicbox=#1}

\usepackage[table]{xcolor}

\title{Are Image Generators Zero-Shot Perceivers? A Rigorous Evaluation}

\addauthor{Shangzhe Di}{dishangzhe@sjtu.edu.cn}{1}
\addauthor{Zhaokai Wang}{wangzhaokai@sjtu.edu.cn}{1}
\addauthor{Weidi Xie}{weidi@sjtu.edu.cn}{1}

\addinstitution{
  Shanghai Jiao Tong University, China
}

\runninghead{Di, Wang, Xie}{Are Image Generators Zero-Shot Perceivers?}

\def\eg{\emph{e.g}\bmvaOneDot}

\usepackage{xcolor}%

\begin{document}

\maketitle

\input{sec/0_abstract}
\input{sec/1_intro}
\input{sec/2_related}
\input{sec/3_method}
\input{sec/4_experiments}
\input{sec/5_conclusion}

\newpage
\bibliography{egbib}

\input{sec/X_supp}

\end{document}

%% file: preamble.tex
\usepackage[dvipsnames]{xcolor}
\usepackage{amsfonts}
\usepackage{amsmath}
\usepackage{amssymb}
\usepackage{dsfont}

\DeclareMathOperator*{\argmin}{arg\,min}

\definecolor{mygreen}{RGB}{29, 177, 0}
\definecolor{darkgreen}{rgb}{0.0, 0.5, 0.0}



%% file: sec/0_abstract.tex
\begin{abstract}
  Recent work, such as Vision Banana, shows that lightweight instruction tuning can enable an image generator to achieve state-of-the-art performance across multiple visual perception tasks.
  Motivated by this perspective, we ask how far image generators can go on public visual perception benchmarks in a zero-shot setting.
  We introduce \textbf{\textsc{ProbeGen}}, a benchmark for \textbf{zero-shot generative perception} that casts monocular depth estimation, referring/reasoning segmentation, and object counting as conditional generation tasks specified through text prompts, and compares 20 models in total---including proprietary and open-weight image generators, specialist perception models, and MLLMs---across 11 published benchmarks.
  We observe that pretrained image generators show measurable zero-shot perceptual competence, but with a clear trade-off: specialist models remain stronger for in-distribution accuracy and efficiency, while generative models are often more robust under distribution shift and better at compositional semantic reasoning.
  We hope this study helps establish zero-shot generative perception as a meaningful research direction and provides a useful foundation for future work at the intersection of visual generation and understanding.

\end{abstract}

%% file: sec/1_intro.tex
\section{Introduction}
\label{sec:intro}

Modern image generators must internalize a rich understanding of the visual world, for example, object structure, geometry, material appearance, and semantic relationships, in order to synthesize realistic scenes~\cite{nano_banana_pro,nano_banana_2,seedream_5_0,gpt_image_2}. This observation invites a provocative hypothesis: 
{the same internal representations that enable generation may already encode the information needed for visual perception.}
Recent work lends credibility to this idea. Vision Banana demonstrates that instruction-tuned generators can be recast as generalist vision learners across a broad task spectrum~\cite{image_generators_generalist_vision}, and video generation models have been shown to exhibit emergent zero-shot reasoning~\cite{video_zero_shot_reasoners}. Yet a fundamental question remains open: when evaluated rigorously on public benchmarks, {how competitive are mainstream image generators as \textit{zero-shot perceivers} compared to state-of-the-art specialist models?}

The current landscape offers partial answers from opposite ends. On the discriminative side, dedicated architectures set strong standards: the Depth Anything series~\cite{da,da2,da3} for monocular depth, the Segment Anything family~\cite{sam,sam2,sam3} for segmentation, and CountGD~\cite{countgd,countgd++} for object counting, {\em etc.} On the generative side, methods such as Marigold \cite{marigold} and Lotus-2~\cite{lotus2} show that diffusion models can be turned into competitive dense predictors, but only after task-specific fine-tuning or architectural modification. Between these two poles lies an unexplored regime that we call \emph{zero-shot generative perception}: the perceptual capability of general-purpose image generators~\cite{nano_banana_pro,nano_banana_2,seedream_5_0,gpt_image_2,flux-2,qwen-image}, evaluated without any adaptation, on the same benchmarks used to measure specialist progress. 

\input{figures/fig_good_case}

We conduct a systematic, controlled study of zero-shot generative perception---which we release as \textsc{ProbeGen}---on three complementary visual perception tasks:
monocular depth estimation, which probes geometric understanding; 
referring/reasoning segmentation, which probes semantic grounding and compositional reasoning; 
and object counting, which probes instance-level localization. 
All three tasks are cast as conditional image generation problems specified entirely through text prompts, yielding a unified inference protocol that accommodates proprietary and open-source generative models.
We evaluate a diverse set of mainstream image generators, \eg, Nano Banana~\cite{nano_banana_pro,nano_banana_2}, Seedream~\cite{seedream_4_5,seedream_5_0}, GPT-Image~\cite{GPT-Image-1,gpt_image_1_5,gpt_image_2}, Flux.2-dev~\cite{flux-2}, and Qwen-Edit-2511~\cite{qwen-image}, 
and benchmark them head-to-head against leading specialist models under identical data and metric conditions.

Our findings reveal that, though specialist models retain a clear advantage in in-distribution accuracy and computational efficiency, general-purpose generators demonstrate markedly greater robustness under distribution shift and superior compositional reasoning on complex queries. 
The gap is most pronounced in out-of-distribution (OOD) domains, for example, manga imagery (Manga109~\cite{manga}) and artistic paintings (DRAM~\cite{DRAM}) in Figure~\ref{fig:good_case}, where the broad visual prior of a generative model provides semantic grounding that specialist pipelines lack.
To better understand this behaviour, we further characterize the systematic failure modes of generative perception---chiefly a failure to preserve the input (regenerating the scene instead of transforming it in place) and a failure to follow the task instruction---that distinguish generative errors from those of conventional discriminative models and point toward concrete avenues for improvement.

In this paper, we make the following contributions:
\textbf{(i)}~we present \textsc{ProbeGen}, the first controlled, benchmark-level comparison (to our knowledge) of mainstream image generators against strong specialist and MLLM baselines for zero-shot visual perception, spanning geometric, semantic, and instance-localization tasks;
\textbf{(ii)}~we design a unified evaluation pipeline that turns raw generator outputs into benchmark-comparable predictions through deterministic extractors applied identically across all models, isolating perceptual ability from post-processing;
\textbf{(iii)}~we identify a consistent accuracy-robustness trade-off: generators underperform specialists in-distribution but match or surpass them under domain shift and compositional complexity;
\textbf{(iv)}~we characterize the systematic failure modes of generative perception, providing concrete diagnostics for future work.

%% file: figures/fig_good_case.tex
\begin{figure}[t]
  \centering
  \includegraphics[width=\linewidth]{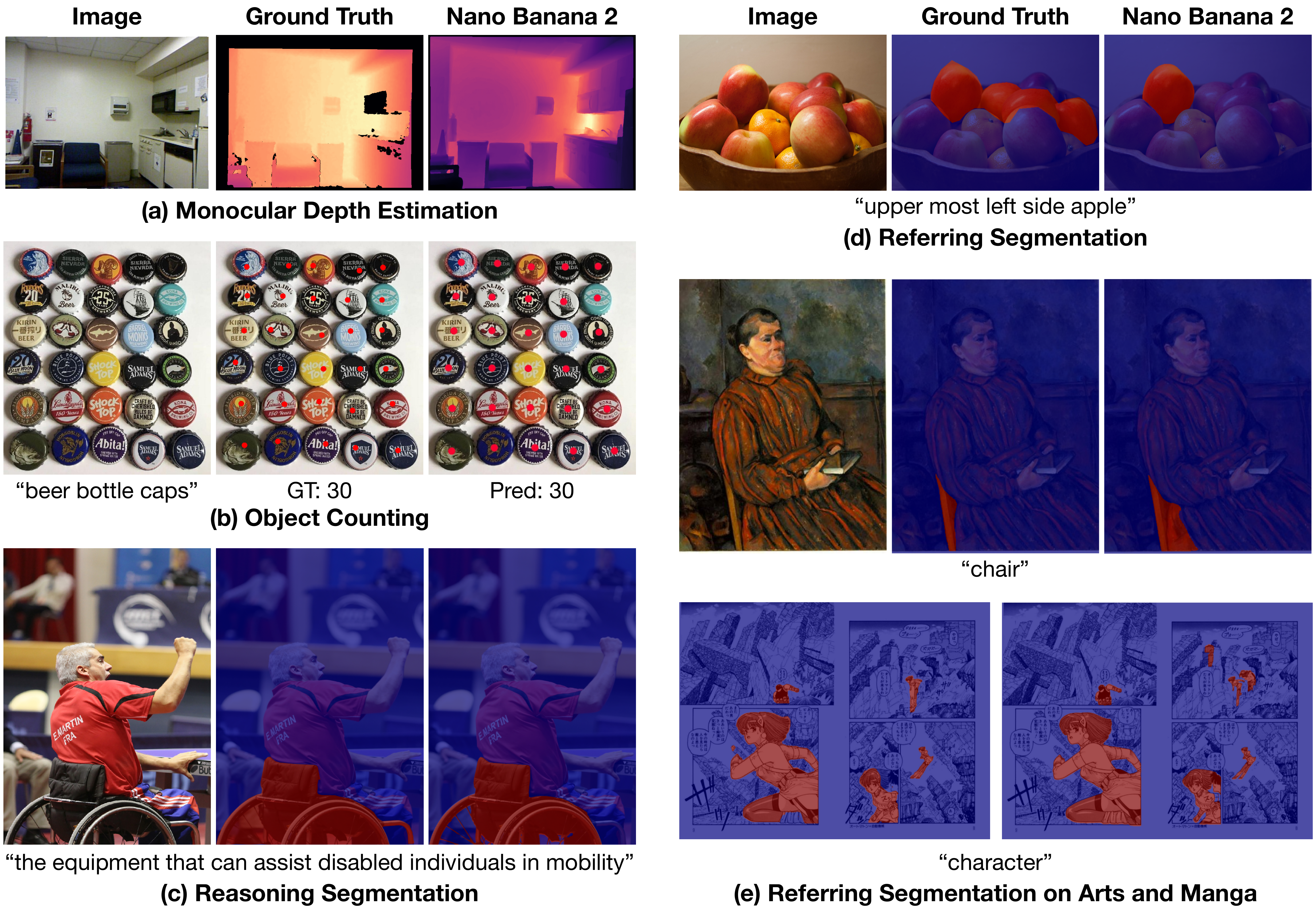}
  \caption{
    \textbf{Zero-shot perception abilities of Nano Banana 2.}
    (a) Monocular depth estimation.
    (b) Object counting.
    (c) Reasoning segmentation.
    (d) Referring segmentation. Nano Banana 2 manages to segment the desired instance despite the incorrect annotation.
    (e) Referring segmentation on out-of-distribution domains of artistic and manga-style images. Note that segmentation results are recoloured on the input image for visualization.
  }\label{fig:good_case}
\end{figure}

%% file: sec/2_related.tex
\section{Related Work}
\label{sec:related}

\noindent \textbf{Discriminative specialists for visual perception.}
Visual perception has traditionally been studied through specialized discriminative models.
In geometry, the Depth Anything series attains strong accuracy and efficiency via task-specific architectures and supervision~\cite{da,da2,da3}.
In semantic localization and segmentation, the Segment Anything line has substantially advanced generic segmentation and open-vocabulary localization~\cite{sam3}.
In object counting, specialist models~\cite{countgd,countgd++,liu2026count} efficiently localize up to hundreds of queried objects.
Such systems define the dominant paradigm for visual perception: each is explicitly built around a particular output space, annotation format, and inference pipeline, and they remain the strongest reference points for in-distribution accuracy and efficiency.
Their success, however, raises a complementary question that motivates our work: can perceptual competence also emerge from generative visual pretraining?

\vspace{5pt}
\noindent \textbf{Adapting image generators into perception specialists.}
This question has become increasingly relevant as a growing body of work explores the connection between image generation and dense perception.
For example, Marigold and Lotus-2 show that diffusion models can become strong dense predictors after task-specific training, architectural modification, or specialized inference design~\cite{marigold,lotus2}.
These results suggest that generative models contain useful perceptual priors, especially for geometric reasoning.
However, they primarily study how to transform a generator into a perception model, rather than whether an image generator can already perform perception tasks out of the box using only its pretrained capabilities.

\vspace{5pt}
\noindent \textbf{Evaluation gaps in zero-shot generative perception.}
Recent studies suggest generative models may support visual understanding beyond synthesis, but important evaluation gaps remain.
Work on video generators reports zero-shot learning and reasoning abilities, yet omits comparison against specialist baselines on public benchmarks, making it difficult to judge how competitive generative models actually are~\cite{video_zero_shot_reasoners}.
Meanwhile, Vision Banana reports promising perception results after instruction tuning, but evaluates only Nano Banana Pro and does not report zero-shot performance~\cite{image_generators_generalist_vision}.
As a result, the native perceptual ability of off-the-shelf image generators remains unclear, motivating the systematic zero-shot evaluation we present here.

%% file: sec/3_method.tex
\section{Zero-Shot Generative Perception}
\label{sec:method}

We formalize the use of pretrained image generators as zero-shot visual perceivers and describe the unified evaluation protocol behind \textsc{ProbeGen}, used throughout this paper.

\vspace{3pt} \noindent \textbf{Disclaimer on ``zero-shot''.} We use \emph{zero-shot} in an operational sense: no task-specific fine-tuning, adaptation, or architectural change, with a text prompt as the only interface to the generator.
This does not guarantee that a generator never saw related supervision during its own pretraining or instruction tuning---for proprietary models, this is undisclosed and unverifiable---so our claims concern zero-shot transfer under a fixed, adaptation-free protocol, and we explicitly flag any model whose training data is known to overlap an evaluated task.
This concern also motivates our OOD segmentation benchmarks (Figure~\ref{fig:good_case}(e)): even if a generator has seen manga (Manga109~\cite{manga}) or paintings (DRAM~\cite{DRAM}) images during training, dense segmentation supervision in these stylized domains is at most a marginal fraction of any plausible training mix---image generators are trained predominantly on image--text pairs for synthesis rather than on segmentation masks---so strong performance there is harder to attribute to data leakage and gives cleaner evidence of genuine zero-shot generalization.

\vspace{3pt} \noindent \textbf{Formulation.}
Let $x \in \mathbb{R}^{H \times W \times 3}$ denote an input RGB image and let $p_t$ denote a task-describing text prompt. A pretrained image generator $G$ produces a task-conditioned output $\hat{y} = G(x, p_t)$, where $\hat{y}$ is itself an RGB image whose pixel content encodes the desired perceptual quantity ({\em e.g.}, a depth map or a segmentation mask rendered as an image). A deterministic, task-specific extraction function $E_t$ then projects $\hat{y}$ into the metric-compatible evaluation space:
\begin{equation}
    \hat{z} = E_t\bigl(\hat{y}\bigr) = E_t\bigl(G(x,\, p_t)\bigr).
    \label{eq:pipeline}
\end{equation}
This decomposition isolates the generator's perceptual capacity, encoded entirely in $\hat{y}$, from the mechanical mapping into a benchmark-compatible format performed by $E_t$. Because $E_t$ is deterministic and fixed identically across all models, all performance differences we report reflect genuine differences in generative perception, not per-model post-processing design. The remainder of this section describes the prompt protocol, the segmentation extractor, the depth extractor, and the counting extractor.

\begin{figure}[t]
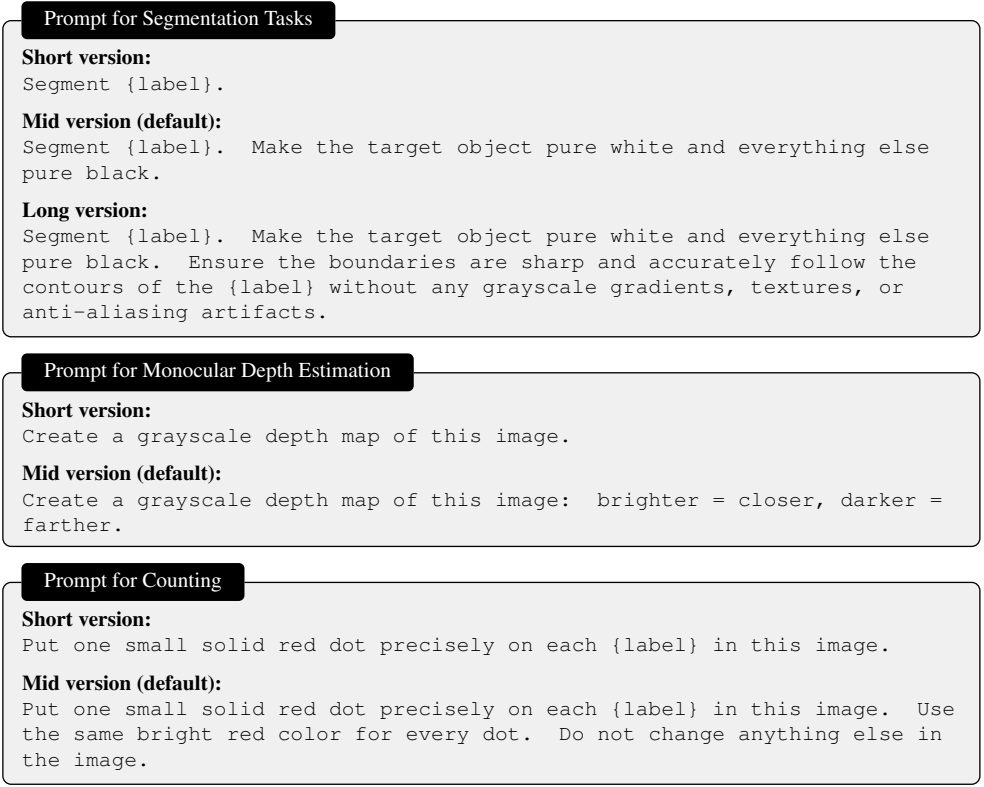

  \begin{AcademicBox}[\footnotesize Prompt for Segmentation Tasks]
    \footnotesize
    \textbf{Short version:}\\
    \texttt{Segment \{label\}.}

    \vspace{5pt}
    \textbf{Mid version (default):}\\
    \texttt{Segment \{label\}. Make the target object pure white and everything else pure black.}

    \vspace{5pt}
    \textbf{Long version:}\\
    \texttt{Segment \{label\}. Make the target object pure white and everything else pure black. Ensure the boundaries are sharp and accurately follow the contours of the \{label\} without any grayscale gradients, textures, or anti-aliasing artifacts.}
  \end{AcademicBox}
  \begin{AcademicBox}[\footnotesize Prompt for Monocular Depth Estimation]
    \footnotesize
    \textbf{Short version:}\\
    \texttt{Create a grayscale depth map of this image.}

    \vspace{5pt}
    \textbf{Mid version (default):}\\
    \texttt{Create a grayscale depth map of this image: brighter = closer, darker = farther.}
  \end{AcademicBox}
  \begin{AcademicBox}[\footnotesize Prompt for Counting]
    \footnotesize
    \textbf{Short version:}\\
    \texttt{Put one small solid red dot precisely on each \{label\} in this image.}

    \vspace{5pt}
    \textbf{Mid version (default):}\\
    \texttt{Put one small solid red dot precisely on each \{label\} in this image. Use the same bright red color for every dot. Do not change anything else in the image.}
  \end{AcademicBox}
  \caption{\textbf{Prompt templates used in our evaluation protocol}.}
  \label{fig:prompt}
\end{figure}

\subsection{Prompt Protocol}

Prompting is the sole task interface for all evaluated generators. Figure~\ref{fig:prompt} summarizes the prompt families. For segmentation, prompts request a greyscale mask in which the target region is white and the background is black. For depth estimation, prompts request a greyscale depth map in which brighter pixels indicate points closer to the camera. For counting, prompts request adding red dots on each target object in the image.

To study how strongly each model depends on linguistic specificity, we design prompts at three granularity levels for segmentation (short, mid, long), two for depth (short, mid), and two for counting (short, mid). The short segmentation prompt provides only the target label; the mid prompt additionally specifies the white-foreground/black-background convention; and the long prompt further requests sharp boundaries without anti-aliasing artefacts. For depth, the short prompt requests a greyscale depth map without specifying polarity, while the mid prompt explicitly states the brighter-is-closer convention. For counting, the short prompt demands putting a red dot on each object, while the mid version further emphasizes an identical dot color and consistency. By default we use the mid versions, and other versions are used only in ablations (Table~\ref{tab:ablate_prompt}).

\subsection{Segmentation Output Extraction}

For referring segmentation and reasoning segmentation, the generator is asked to produce a high-contrast mask corresponding to the queried target. Because raw outputs are continuous RGB images rather than discrete label maps, we apply a deterministic binarization step before computing segmentation metrics.

\vspace{3pt} \noindent \textbf{Extraction strategy.} 
We benchmark four deterministic binarization strategies, each applied identically across all models, to ensure that our conclusions are not artefacts of a particular extraction choice:
\begin{equation}
    E_\text{seg}(\hat{y}) = \mathds{1} \bigl[\,\text{gray}(\hat{y}) > \tau\,\bigr],
\end{equation}
where $\text{gray}(\cdot)$ converts to single-channel luminance and $\tau$ is determined by one of the following thresholds: (i)~a fixed intensity threshold at 128, 
(ii)~a fixed threshold at 225, (iii)~K-means clustering with $k{=}2$, and (iv)~Otsu's adaptive thresholding~\cite{otsu1979threshold}, which selects the binarization boundary by maximizing inter-class variance. Based on the ablation study in Table~\ref{tab:ablate_mask}, Otsu thresholding yields the most robust binarization and consistently suppresses the soft-boundary artefacts characteristic of raw generative outputs across six segmentation benchmarks. We therefore adopt Otsu thresholding as the default extractor in all main segmentation experiments.

\subsection{Monocular Depth Extraction}
\label{sec:method-depth}

For depth estimation, each generator is prompted to synthesize a greyscale depth map from a single RGB image. Unlike specialist depth estimators that directly predict metrically calibrated depth, general image generators typically output relative depth maps in which pixel intensity reflects proximity but absolute scale and offset are unconstrained.

\vspace{3pt} \noindent \textbf{Affine-invariant evaluation.} 
We therefore evaluate geometric understanding under an affine-invariant protocol following standard practice~\cite{marigold,midas}. Specifically, we request outputs where brighter pixels denote regions closer to the camera, and then align the predicted map $\hat{d} = \text{gray}(\hat{y})$ to ground truth $d$ via optimal scale-and-shift fitting:
\begin{equation}
    (s^*, t^*) = \argmin_{s,\,t} \|s\hat{d} + t\mathbf{1} - d\|_2^2,
\end{equation}
where the scalar scale $s$ and shift $t$ admit a closed-form least-squares solution, and all metrics are computed on the aligned prediction $\tilde{d} = s^* \hat{d} + t^*$.
This post-hoc alignment removes the arbitrary linear ambiguity inherent to generative outputs while preserving the structural and ordinal information encoded in the generated depth map.

\subsection{Object Counting Extraction}
\label{sec:method-counting}

Unlike depth and segmentation, counting only requires a single integer per image rather than a dense map. We probe this capability through a \textit{mark-then-count} interface: as in Figure~\ref{fig:prompt}, the generator is prompted to paint one small solid red dot on every instance of the target category while leaving the rest of the image unchanged. The marked image $\hat{y}$ is then decoded into an instance count by a deterministic extractor $E_\text{count}$ that recovers how many valid dots the model deposited.

\vspace{3pt} \noindent \textbf{Three-stage decoder.}
$E_\text{count}$ operates at the preprocessed resolution (long side 1024) so that the comparison against the original conditioning image is pixel-aligned: (i) \textit{colour isolation} keeps only saturated red pixels via two HSV ranges that straddle the hue wrap around $0/180$; (ii) a \textit{diff gate} retains only red pixels that changed relative to the input, suppressing red structures already present in the scene ({\em e.g.}, a stop sign or red packaging); and (iii) a \textit{shape-filtered blob detector} returns a set of keypoints whose cardinality is the predicted count $\hat{c}$, where each candidate must satisfy fixed area, circularity, and inertia-ratio constraints. The shape filter is what separates true markers from elongated red streaks or oversized red regions that survive the colour and diff gates. All hyperparameters are fixed across all generators by a leakage-safe tuning protocol; details are deferred to Appendix~\ref{sec:supp-counting}.

%% file: sec/4_experiments.tex
\section{Experimental Results}
\label{sec:experiments}

\subsection{Setup}
\label{sec:exp_setup}

\noindent \textbf{Benchmarks.}
We evaluate \textsc{ProbeGen} across diverse tasks and domains.
For monocular depth estimation, we employ NYUv2~\cite{nyuv2}, DIODE~\cite{diode}, ScanNet~\cite{scannet}, and KITTI~\cite{kitti} as standard benchmarks and report the AbsRel metric.
For referring segmentation, we report results on three RefCOCO variants~\cite{refcoco,refcocog}.
For reasoning segmentation, we adopt ReasonSeg~\cite{lisa}.
To assess robustness in stylized domains, we additionally evaluate on Manga109~\cite{manga} (manga segmentation) and DRAM~\cite{DRAM} (art paintings), which probe perception beyond photorealistic imagery. 
We report gIoU and cIoU metrics for all of the segmentation benchmarks.
For counting, we use the PixMo-Points dataset~\cite{pixmo} and report MAE and RMSE metrics.
Due to constraints on evaluation time and API cost, we randomly sample 500 examples per benchmark, for a total of 5,500 samples across the 11 benchmarks.

\vspace{5pt}
\noindent \textbf{Baselines.} We compare three baseline families:
(1) \textit{Specialists}, including Depth Anything V2~\cite{da2} and the task-specific diffusion models Lotus-2~\cite{lotus2} and Marigold~\cite{marigold} for depth, SAM3 Agent~\cite{sam3,qwen3-vl} and Qwen3-VL-SAMTok~\cite{samtok} for segmentation, and CountGD++~\cite{countgd++} for counting;
(2) \textit{Generative Generalists}, namely general-purpose image generators evaluated in a zero-shot manner across all perception tasks;
and (3) \textit{MLLMs}, namely general-purpose vision-language models such as Gemini-3.5-Flash~\cite{gemini_3_5}, GPT-5.4~\cite{gpt_5_4}, and Claude-Opus-4.7~\cite{claude_opus_4_7} that read the input image and emit the answer directly as text rather than synthesizing an output image; since their output is a scalar rather than a dense pixel map, we evaluate them on counting only.
Since Vision Banana is not publicly accessible~\cite{image_generators_generalist_vision}, we cite its reported numbers for reference only; they are not directly comparable, as they use the full benchmark rather than our 500-image subset, with different prompts and post-processing.

\vspace{5pt}
\noindent \textbf{Inference.} All generative models are evaluated with a single sample per image under deterministic decoding (if possible). Inputs and outputs are standardized to a long side of 1024 while preserving the aspect ratio.
Unless otherwise noted, all results follow the default post-processing in Sec.~\ref{sec:method}: segmentation results with Otsu-based mask extraction, depth results with affine-invariant alignment, and counting results with the three-stage decoder.

\vspace{5pt}
\noindent \textbf{Efficiency Evaluation.}
Locally deployed models are benchmarked on a single NVIDIA H800 (batch size 1, long side 1024, after warm-up), reporting throughput (IMG/min) and peak GPU memory. API-hosted proprietary models expose neither, so their efficiency is excluded from the main tables and reported for reference only in Appendix~\ref{sec:supp-efficiency}.

\vspace{5pt}
\noindent \textbf{Notation.}
Across all result tables, $\times$ denotes a failure to produce a valid output for the task, {--} an unavailable metric, and $^\ddag$ Vision Banana numbers reported under a different protocol (full benchmark, with different prompts and post-processing). \textbf{Bold numbers} indicate the best performance within the \textit{Generative Generalists} group.

\input{tables/reason_and_ood_seg}
\input{tables/ref_seg}
\input{tables/depth}

\subsection{Referring and Reasoning Segmentation}

\textbf{Generative models excel at reasoning and OOD segmentation.} 
General image generators are particularly competitive on reasoning-heavy and OOD segmentation, as reported in Table~\ref{tab:reason_ood_seg}. On ReasonSeg, Nano Banana 2 achieves the strongest result among all directly comparable systems, reaching 70.3 gIoU and outperforming the multi-step agentic SAM3 + Qwen3-VL-8B-Thinking pipeline (60.7 gIoU). This result suggests that, for compositional segmentation queries, a single generative forward pass can be more effective than explicit tool-calling pipelines. 
The advantage of generative models is even clearer in OOD scenarios. On DRAM and Manga109, Nano Banana Pro and Nano Banana 2 substantially outperform the specialists. Figure~\ref{fig:good_case}(e) uses Nano Banana 2 as a representative strong generator and shows that this robustness extends qualitatively to artistic and manga-style inputs. These results indicate that general image generators can preserve semantic binding more robustly when the visual domain departs from natural-image statistics.
Although the Vision Banana result is not directly comparable ($^\ddag$), its strong ReasonSeg number still suggests that the success of Vision Banana likely depends in large part on the underlying capability of Nano Banana Pro itself, rather than only on task-specific tuning.

\vspace{5pt}
\noindent \textbf{Generators remain competitive in standard referring segmentation.} 
On standard referring segmentation benchmarks in Table~\ref{tab:refseg}, specialist systems still maintain an advantage in overall peak performance, especially when trained or engineered directly for segmentation. Here again, the Vision Banana result$^\ddag$ is broadly consistent with our observation that Nano Banana Pro is already a very capable zero-shot generator for segmentation, suggesting that the gains reported by Vision Banana are built on top of a strong pretrained foundation. Nevertheless, the best general image generators remain highly competitive: Nano Banana Pro reaches 70.7 gIoU on RefCOCO and 72.0 gIoU on RefCOCOg, approaching strong specialist baselines without any task-specific fine-tuning. Figure~\ref{fig:good_case} further uses Nano Banana 2 as a representative example to show that a single general generator can handle both referring and reasoning segmentation in a fully zero-shot manner. Notably, although the ground-truth annotation in the second row of Figure~\ref{fig:good_case}(d) is incorrect, Nano Banana 2 still follows the instruction and segments the correct apple instance.


\vspace{5pt}
\noindent \textbf{Large gap among general generators.} The results also reveal a clear gap within the generative generalist family: Nano Banana models are consistently strong across all segmentation benchmarks, whereas Seedream-4.5 fails to produce usable masks at all, and Flux.2-dev and Qwen-Edit-2511 are substantially weaker, especially on reasoning-heavy segmentation.

\subsection{Monocular Depth Estimation}

\textbf{Specialists still lead in the accuracy-efficiency trade-off.} Table~\ref{tab:depth} reports relative monocular depth estimation under affine-invariant evaluation. The Vision Banana results$^\ddag$ suggest a useful contrast: although Vision Banana applies instruction tuning for perception, its depth performance does not consistently surpass Qwen-Edit-2511 (\eg, on NYUv2 and KITTI), indicating that depth benefits more from models with stronger depth-specific training or supervision than from general perception-oriented tuning alone. Specialist models remain clearly superior in both accuracy and efficiency: Depth Anything V2-Large achieves strong accuracy while running at 828 images per minute, and the task-specific generative model Lotus-2 reaches the same level of accuracy with substantially lower throughput.

\vspace{5pt}
\noindent \textbf{Depth exposure of Qwen-Edit.} At the same time, several general image generators exhibit non-trivial zero-shot geometric competence. Among them, Qwen-Edit-2511 is the strongest open-weight general generator in our evaluation, achieving 5.6 AbsRel on NYUv2, 17.1 on DIODE, 6.5 on ScanNet, and 8.9 on KITTI; on DIODE and KITTI it even surpasses the specialist baselines. This result, however, must be interpreted with care: the Qwen-Image technical report states that Qwen-Edit-2511's instruction-tuning data includes depth-estimation examples~\cite{qwen-image}, so depth is effectively in-distribution for this model. Under our operational definition (Sec.~\ref{sec:method}), it still satisfies the adaptation-free, prompt-only protocol---we add no fine-tuning and prompt it exactly as we do every other model---but its depth scores reflect prior task exposure and are therefore best read as a near-supervised reference point rather than evidence of emergent geometry. We flag this case explicitly because the training corpora of the proprietary generators are undisclosed, so we likewise cannot rule out analogous exposure for them; our zero-shot claims thus concern transfer under a fixed protocol, not provable novelty of each task to each model.

\vspace{5pt}
\noindent \textbf{Comparison among generative models.} Nano Banana Pro and Nano Banana 2 remain noticeably weaker in absolute depth accuracy, but still produce coherent relative geometry without any depth-specific adaptation on our part. Figure~\ref{fig:good_case} uses Nano Banana 2 as a representative example and shows that one model can jointly support depth estimation and segmentation through the same zero-shot interface. In contrast, other general image generators either degrade substantially or fail outright on this task, as seen for Seedream-5.0 and Flux.2-dev. Overall, these results suggest that relative scene geometry can emerge in general image generators, but accurate and efficient depth prediction still strongly favours specialist architectures or task-adapted generative models.

\subsection{Object Counting}
\input{tables/counting}

\textbf{MLLMs lead counting in MAE.} Table~\ref{tab:counting} reports object counting results on PixMo-Points~\cite{pixmo}. Unlike depth and segmentation, counting reveals a markedly different ordering across baseline families. The strongest results come from \textit{multimodal LLMs} that emit the count as a single integer: Gemini-3.5-Flash~\cite{gemini_3_5} reaches MAE 4.4, followed by Claude-Opus-4.7~\cite{claude_opus_4_7} at 6.7 and GPT-5.4~\cite{gpt_5_4} at 8.6.
Notably, their advantage is confined to MAE: on RMSE, the best MLLM (19.9) is worse than the best generators (15.5--15.6), suggesting that MLLMs occasionally make large errors on densely populated scenes, whereas the mark-then-count interface degrades more gracefully.
Apart from being exposed to massive counting-related data during pre-training, MLLMs adopt a native numerical output paradigm well-suited for counting, different from the dense prediction paradigm in depth and segmentation. MLLMs directly predict the final number without requiring intermediate marking of each object.

\vspace{5pt}
\noindent \textbf{Degradation of the counting specialist.} CountGD++~\cite{countgd++} trails at MAE 16.7 despite a strong throughput of 471~IMG/min (Table~\ref{tab:supp-efficiency}), which we attribute to a mismatch between its training distribution and the long-tail, open-vocabulary categories in PixMo-Points.

\vspace{5pt}
\noindent \textbf{General image generators sit between two extremes.} Nano Banana 2 is the strongest generator at MAE~7.3 / RMSE~15.5, with Nano Banana Pro and GPT-Image-2 close behind. This is the same Nano-Banana-dominance pattern that emerges in our segmentation results, suggesting a model-family property rather than a task-specific advantage. The remaining generators, Seedream-4.5/5.0~\cite{seedream_4_5,seedream_5_0}, Flux.2-dev~\cite{flux-2}, and Qwen-Edit-2511~\cite{qwen-image}, are unable to follow the dot-marking instruction and collapse to invalid outputs. This is consistent with their behaviour on reasoning segmentation, where the same models also struggle with the format-discipline aspect of the task.

\vspace{5pt}
\noindent \textbf{Accuracy-efficiency trade-off across interfaces, not just across model families.} Emitting an integer is more accurate on average (MAE) and substantially cheaper than rendering an annotated image, while image-level marking yields outputs that are visually interpretable, more robust in the worst case (RMSE), but bounded by what the generator is willing to overlay on the scene. Two residual error sources are inherent to the mark-then-count probe: red-cluttered scenes (where the generator paints into a background that already contains red structures) and densely packed instances (where adjacent dots merge into a single blob). Both are generation-side limitations that no post-processing pipeline can fully recover.

\subsection{Ablation Studies}
\label{sec:ablation}

To limit API cost, all ablations are conducted on a 100-image subset of each benchmark; absolute numbers therefore differ, but the relative comparisons are unaffected.

\vspace{5pt}
\noindent \textbf{Choice of mask extraction method.}
Table~\ref{tab:ablate_mask} validates the segmentation extraction procedure on Nano Banana 2. The Otsu thresholding achieves the best average performance across the six segmentation benchmarks, yielding 68.8 average gIoU and outperforming fixed thresholds and K-means clustering. This supports our choice of Otsu thresholding as the default binarization strategy in the main experiments.

\vspace{5pt}
\noindent \textbf{Sensitivity to prompt granularity.} Table~\ref{tab:ablate_prompt} shows that a model's sensitivity to prompt wording is inversely related to its capability. The Nano Banana series is essentially flat across all prompt variants (\eg, Nano Banana 2 scores 77.0/76.7/75.7 gIoU on RefCOCOg), so its strength is not an artefact of careful prompt engineering; the most verbose prompt even hurts slightly, indicating that the mid prompt is already near-optimal. Weaker generators are far more prompt-dependent. On segmentation, where three granularities are available, their gains are concentrated almost entirely at the short$\to$mid step: GPT-Image-2 jumps from 24.6 to 52.5 gIoU and Seedream-5.0 from 8.0 to 40.1 once the mid prompt states the white-foreground/black-background convention. The following mid$\to$long step, which only asks for sharper boundaries, adds comparatively little (GPT-Image-2 $52.5\to57.9$) and can even hurt (Seedream-5.0 $40.1\to37.5$, Qwen-Edit-2511 $36.3\to34.2$). For these models the real bottleneck is thus specifying the output \emph{format}, not refining the boundary language. Depth and counting use only short and mid prompts, and there the same short$\to$mid improvement appears but is much smaller---adding brighter-is-closer polarity for depth (GPT-Image-2 $10.9\to9.6$ AbsRel) or a uniform-dot-colour constraint for counting (Nano Banana Pro $5.1\to4.7$ MAE) helps only marginally for models that already succeed. Overall, the mid prompt is a fair, near-optimal default for every model, so our main conclusions do not depend on per-model prompt tuning.

\input{tables/ablate_mask}
\input{tables/ablate_prompt}

\subsection{Failure Modes}
\label{sec:failure}

\textbf{Alignment and format cause failures.} We highlight \textit{systematic, model-level failures}, where some generators broadly lack the capability required for a given visual perception task rather than merely failing on a few difficult examples. As shown in Figure~\ref{fig:bad_case1}, these failures mainly arise from two sources. The first is \textit{failure to preserve the input}, where the generator regenerates the scene instead of transforming it in place, so the output is no longer registered to the input image. This spans a range of severity, from geometric drift that breaks pixel-wise correspondence (Figure~\ref{fig:bad_case1}(b)) to wholesale re-rendering that hallucinates new content, such as the redrawn bottle-cap designs in Figure~\ref{fig:bad_case1}(c). The second is an \textit{instruction-following failure}, where the model does not perform the requested transformation at all, returning a stylized or near-unchanged copy of the input instead of a valid binary mask, depth map, or marked image (\eg, the Seedream outputs in Figure~\ref{fig:bad_case1}(a,b)). Consistent with this, such models improve sharply once the prompt explicitly states the output format (Table~\ref{tab:ablate_prompt}). This distinction helps separate models that fundamentally cannot perform a task from models that can perform it but still fail in challenging instances.

\vspace{5pt}
\noindent \textbf{Unified understanding-and-generation models do not yet yield emergent perception.} We also tested BAGEL~\cite{bagel}, OmniGen2~\cite{omnigen2}, and Lance~\cite{lance}, which couple understanding and generation natively, and UNO~\cite{uno}, an in-context Flux variant. Under every prompt granularity, none produced a usable mask, depth map, or dot annotation on any benchmark; they return the input with only a global, filter-like alteration, {\em i.e.}, the instruction-following failure above in its purest form, so we omit them from the tables. The bottleneck is instruction-following for pixel-registered, structured outputs---precisely what lightweight instruction tuning (as in Vision Banana) supplies.

\vspace{5pt}
\noindent \textbf{Surface normals remain unsolved for zero-shot settings.} We also explored surface normal estimation with generative models, but found that despite the success of Vision Banana in this task, none of the evaluated image generators could complete it reliably in zero-shot settings. As shown in Figure~\ref{fig:bad_case3}, we instructed the model to output an RGB image whose three channels encode the X, Y, and Z components of the normal vector, respectively. Some models, such as Nano Banana and GPT-Image, could produce visually plausible normal-like maps, but the predictions were entirely inaccurate. This failure may stem from incomplete adherence to the instruction-defined RGB-to-normal mapping, or simply from the scarcity of surface-normal supervision in the models' training data.


\input{figures/fig_bad_case1}
\input{figures/fig_bad_case3}

\subsection{Discussion}

\textbf{Three ingredients of zero-shot generative perception.} Taken together, our results suggest that zero-shot generative perception rests on three ingredients, and a weakness in any one of them is sufficient to cause failure. \textit{Semantic grounding} determines whether the generator identifies the right target; it is the ingredient where generators are strongest, and it explains their advantage on reasoning-heavy and OOD segmentation. \textit{Spatial faithfulness} requires the output to remain pixel-registered to the input; its absence is the input-preservation failure of Sec.~\ref{sec:failure}. For relative depth, a global scale-and-shift alignment can absorb much of this error, but for masks, surface normals, and counting the misalignment is local and no deterministic extractor can undo it. \textit{Output-format discipline} requires the generator to emit a strict mask, depth map, or marked image rather than a stylized edit; its absence is the instruction-following failure of Sec.~\ref{sec:failure}, and it is the ingredient that lightweight instruction tuning most directly addresses. Progress on the latter two will likely require, beyond perception supervision, explicit spatial-consistency objectives and reward signals that penalize structural misalignment.

\vspace{5pt}
\noindent \textbf{Efficiency remains a major issue.} Even the strongest generators run one to three orders of magnitude slower than specialist models and require substantially more GPU memory. Full efficiency measurements are reported in Table~\ref{tab:supp-efficiency} (Appendix~\ref{sec:supp-efficiency}).

\vspace{5pt}
\noindent \textbf{Generative perception remains promising.} Despite these limitations, general image generation models remain attractive for perception because a single interface can already support segmentation, depth estimation, and other structured tasks without task-specific heads. This flexibility is especially promising for open-world concepts, long-tail queries, and natural-language interaction. Although current models are still inefficient and imperfect, their zero-shot transfer and cross-domain robustness suggest substantial long-term potential if alignment, faithfulness, and efficiency can be improved together. We discuss limitations and concrete extensions of \textsc{ProbeGen} in Appendix~\ref{sec:supp-limitations}.

%% file: tables/reason_and_ood_seg.tex
\begin{table*}[t]
  \centering
  \begin{threeparttable}
    
    \resizebox{\linewidth}{!}{%
    \begin{tabular}{l cc: cc cc: cc}
      \toprule
      \multirow{2}{*}{\textbf{Model}} & \multicolumn{2}{c}{\textbf{ReasonSeg}} & \multicolumn{2}{c}{\textbf{DRAM}} & \multicolumn{2}{c}{\textbf{Manga109}} & \textbf{Speed} & \textbf{GPU Mem} \\
      \cmidrule(lr){2-3} \cmidrule(lr){4-5} \cmidrule(lr){6-7} \cmidrule(lr){8-8} \cmidrule(lr){9-9}
      & gIoU & cIoU & gIoU & cIoU & gIoU & cIoU & IMG/min & GB \\
      \midrule
      \textcolor{gray}{\textit{Specialists}} &&&&&&&& \\
      \hspace{0.5em} SAM3 + {Qwen3-VL-8B-Thinking}~\cite{sam3} & 60.7 & 55.3 & 63.5 & 63.1 & 13.7 & 12.4 & 2.6 & 29 \\
      \hspace{0.5em} SAM3 + {Qwen3-VL-32B-Thinking-FP8}~\cite{sam3} & 62.5 & 57.3 & 63.0 & 61.9 & 18.0 & 17.6 & 2.1 & 73 \\
      \hspace{0.5em} Qwen3-VL-8B-SAMTok~\cite{samtok} & 34.3 & 43.4 & 75.5 & 76.5 & 14.5 & 13.5 & 54  & 19 \\

      \midrule
      \textcolor{gray}{\textit{Generative Generalists}} &&&&&&&& \\
      \hspace{0.5em} \textcolor{gray}{Vision Banana$^\ddag$~\cite{image_generators_generalist_vision}} & \textcolor{gray}{79.3} & \textcolor{gray}{{--}} & \textcolor{gray}{{--}} & \textcolor{gray}{{--}} & \textcolor{gray}{{--}} & \textcolor{gray}{{--}} & \textcolor{gray}{{--}} & \textcolor{gray}{{--}} \\
      \hspace{0.5em} Nano Banana Pro~\cite{nano_banana_pro} & 70.2 & 66.1 & 81.8 &  \textbf{81.0} &  \textbf{55.9} &  \textbf{58.8} & {--} & {--} \\
      \hspace{0.5em} Nano Banana 2~\cite{nano_banana_2} &  \textbf{70.3} &  \textbf{68.0} &  \textbf{82.5} & 79.1 & 55.4 & 58.7 & {--} & {--} \\
      \hspace{0.5em} GPT-Image-2~\cite{gpt_image_2} & 41.6 & 42.8 & 64.1 & 69.1 & 28.4 & 29.8 & {--} & {--} \\
      \hspace{0.5em} Seedream-4.5~\cite{seedream_4_5} & {$\times$} & {$\times$} & {$\times$} & {$\times$} & {$\times$} & {$\times$} & {$\times$} & {$\times$} \\
      \hspace{0.5em} Seedream-5.0~\cite{seedream_5_0} & 21.3 & 21.6 & 33.9 & 31.1 & 13.9 & 13.2 & {--} & {--} \\
      \hspace{0.5em} Flux.2-dev-NF4 (32B)~\cite{flux-2} & 21.8 & 19.5 & 49.9 & 46.3 & 9.2 & 8.9 & 1.2 & 39 \\
      \hspace{0.5em} Qwen-Edit-2511 (20B)~\cite{qwen-image} & 25.2 & 25.8 & 49.4 & 38.7 & 13.0 & 9.8 & 1.5 & 61 \\

      \bottomrule
    \end{tabular}
    }

    \vspace{5pt}
    \caption{
      \textbf{Reasoning and OOD Segmentation.}
      We report the gIoU and cIoU metrics.
      GPT-Image 1 and 1.5~\cite{GPT-Image-1,gpt_image_1_5} are not included since they fail to achieve pixel-level consistency (Figure~\ref{fig:bad_case1}). Speed and GPU memory are reported for locally deployed models only; API-hosted models are listed in Table~\ref{tab:supp-efficiency}.
    }
    \label{tab:reason_ood_seg}
    
  \end{threeparttable}
\end{table*}

%% file: tables/ref_seg.tex
\begin{table}[t]
  \centering
  \begin{threeparttable}

    \resizebox{.9\linewidth}{!}{%
    \begin{tabular}{l cSS cSS cSS}
      \toprule
      \multirow{2}{*}{\textbf{Model}} && \multicolumn{2}{c}{\textbf{RefCOCO}} && \multicolumn{2}{c}{\textbf{RefCOCO+}} && \multicolumn{2}{c}{\textbf{RefCOCOg}} \\
      \cmidrule{3-4} \cmidrule{6-7} \cmidrule{9-10}
      && gIoU & cIoU && gIoU & cIoU && gIoU & cIoU \\

      \midrule
      \multicolumn{4}{l}{\textcolor{gray}{\textit{Specialists}}} \\
      \hspace{0.5em} SAM3 + {Qwen3-VL-8B-Thinking}~\cite{sam3}      && 70.7 & 69.3 && 62.1 & 57.6 && 69.8 & 67.2 \\
      \hspace{0.5em} SAM3 + {Qwen3-VL-32B-Thinking-FP8}~\cite{sam3} && 76.1 & 75.8 && 68.6 & 65.5 && 72.3 & 70.7 \\
      \hspace{0.5em} Qwen3-VL-8B-SAMTok*~\cite{samtok}                && 79.4 & 76.8 && 77.7 & 72.8 && 77.6 & 75.4 \\

      \midrule
      \textcolor{gray}{\textit{Generative Generalists}} \\
      \hspace{0.5em} \textcolor{gray}{Vision Banana$^\ddag$~\cite{image_generators_generalist_vision}} && \textcolor{gray}{{--}} & \textcolor{gray}{{--}} && \textcolor{gray}{{--}} & \textcolor{gray}{{--}} && \textcolor{gray}{{--}} & \textcolor{gray}{73.8} \\
      \hspace{0.5em} Nano Banana Pro~\cite{nano_banana_pro}       &&  \textbf{70.7} & \textbf{63.7} && \textbf{60.9} & \textbf{53.6} && \textbf{72.0} &  \textbf{67.1} \\
      \hspace{0.5em} Nano Banana 2~\cite{nano_banana_2}         &&  70.4 & 63.3 && 58.7 & 53.0 && 70.7 & 63.9 \\
      \hspace{0.5em} GPT-Image-2~\cite{gpt_image_2} && 56.3 & 50.0 && 48.0 & 40.3 && 37.9 & 34.7 \\
      \hspace{0.5em} Seedream-4.5~\cite{seedream_4_5}          && {$\times$} & {$\times$} && {$\times$} & {$\times$} && {$\times$} & {$\times$} \\
      \hspace{0.5em} Seedream-5.0~\cite{seedream_5_0}          && 22.9 & 18.7 && 19.9 & 16.1 && 23.3 & 19.1 \\
      \hspace{0.5em} Flux.2-dev-NF4 (32B)~\cite{flux-2}  && 31.1 & 22.6 && 21.1 & 17.3 && 30.9 & 25.1 \\
      \hspace{0.5em} Qwen-Edit-2511 (20B)~\cite{qwen-image}  && 30.9 & 21.6 && 21.8 & 15.7 && 32.0 & 22.8 \\

      \bottomrule
    \end{tabular}
    }
  \end{threeparttable}

  \vspace{5pt}
  \caption{
      \textbf{Referring Segmentation}.
      We report the gIoU and cIoU metrics.
      *: Qwen3-VL-8B-SAMTok has been trained on these benchmarks.
    }
    \label{tab:refseg}
    
\end{table}

%% file: tables/depth.tex
\begin{table*}[t]
  \centering
  \begin{threeparttable}
    
    \resizebox{\linewidth}{!}{%
    \begin{tabular}{l cccc:cc}
      \toprule
      \textbf{Model} & \textbf{NYUv2} & \textbf{DIODE} & \textbf{ScanNet} & \textbf{KITTI} & {\textbf{Speed} (IMG/min)} & {\textbf{GPU Mem} (GB)} \\
      \midrule
      \textcolor{gray}{\textit{Specialists}} &&&&&& \\
      \hspace{0.5em} Depth Anything V2-Large~\cite{da2} & 4.5 & 18.6 & 4.2 & 10.3 & 828 & 3.5 \\
      \hspace{0.5em} Marigold v1.1~\cite{marigold} & 6.0 & 20.1 & 6.8 & 13.7 & 226 & 10 \\
      \hspace{0.5em} Lotus-2~\cite{lotus2} & 4.6 & 18.5 & 4.7 & 10.2 & 25 & 34 \\

      \midrule
      \textcolor{gray}{\textit{Generative Generalists}} &&&&&& \\
      \hspace{0.5em} \textcolor{gray}{Vision Banana$^\ddag$~\cite{image_generators_generalist_vision}} & \textcolor{gray}{8.1} & \textcolor{gray}{10.8} & \textcolor{gray}{{--}} & \textcolor{gray}{10.7} & \textcolor{gray}{{--}} & \textcolor{gray}{{--}} \\
      \hspace{0.5em} Nano Banana Pro~\cite{nano_banana_pro} & 12.3 & 34.9 & 11.8 & 17.7 & {--} & {--} \\
      \hspace{0.5em} Nano Banana 2~\cite{nano_banana_2} & 12.8 & 34.4 & 11.9 & 18.3 & {--} & {--} \\
      \hspace{0.5em} GPT-Image-2~\cite{gpt_image_2} & 11.2 & 29.7 & 10.6 & 16.9 & {--} & {--} \\
      \hspace{0.5em} Seedream-4.5~\cite{seedream_4_5} & 13.1 & 23.2 & 11.8 & 23.4 & {--} & {--} \\
      \hspace{0.5em} Seedream-5.0~\cite{seedream_5_0} & {$\times$} & {$\times$} & {$\times$} & {$\times$} & {$\times$} & {$\times$} \\
      \hspace{0.5em} Flux.2-dev-NF4 (32B)~\cite{flux-2} & {$\times$} & {$\times$} & {$\times$} & {$\times$} & {$\times$} & {$\times$} \\
      \hspace{0.5em} Qwen-Edit-2511 (20B)~\cite{qwen-image} &  \textbf{5.6} &  \textbf{17.1} &  \textbf{6.5} &  \textbf{8.9} & 1.5 & 62 \\

      \bottomrule
    \end{tabular}
    }

    \vspace{5pt}
    \caption{
      \textbf{Monocular Depth Estimation}.
      We report the AbsRel metric (lower is better $\downarrow$). Speed and GPU memory are reported for locally deployed models only; API-hosted models are listed in Table~\ref{tab:supp-efficiency}.
    }
    \label{tab:depth}
    
  \end{threeparttable}
\end{table*}

%% file: tables/counting.tex
\begin{wraptable}{r}{0.5\textwidth}
  \vspace{-12pt}
  \centering
  \begin{threeparttable}
    \resizebox{\linewidth}{!}{%
    \begin{tabular}{l cc}
      \toprule
      \textbf{Model} & \textbf{MAE} $\downarrow$ & \textbf{RMSE} $\downarrow$ \\
      \midrule
      \textcolor{gray}{\textit{Specialists}} && \\
      \hspace{0.5em} CountGD++~\cite{countgd++} & 16.7 & 46.7 \\
      \midrule
      \textcolor{gray}{\textit{MLLMs}} && \\
      \hspace{0.5em} Gemini-3.5-Flash~\cite{gemini_3_5} & 4.4 & 23.7 \\
      \hspace{0.5em} GPT-5.4~\cite{gpt_5_4} & 8.6 & 44.4 \\
      \hspace{0.5em} Claude-Opus-4.7~\cite{claude_opus_4_7} & 6.7 & 19.9 \\
      \midrule
      \textcolor{gray}{\textit{Generative Generalists}} && \\
      \hspace{0.5em} Nano Banana Pro~\cite{nano_banana_pro} & 7.6 & 15.6 \\
      \hspace{0.5em} Nano Banana 2~\cite{nano_banana_2} & \textbf{7.3} & \textbf{15.5} \\
      \hspace{0.5em} GPT-Image-2~\cite{gpt_image_2} & 8.6 & 15.6 \\
      \hspace{0.5em} Seedream-4.5~\cite{seedream_4_5} & {$\times$} & {$\times$} \\
      \hspace{0.5em} Seedream-5.0~\cite{seedream_5_0} & {$\times$} & {$\times$} \\
      \hspace{0.5em} Flux.2-dev-NF4 (32B)~\cite{flux-2} & {$\times$} & {$\times$} \\
      \hspace{0.5em} Qwen-Edit-2511 (20B)~\cite{qwen-image} & {$\times$} & {$\times$} \\

      \bottomrule
    \end{tabular}
    }

    \vspace{5pt}
    \caption{
      \textbf{Object Counting.}
      We report the MAE and RMSE (lower is better $\downarrow$).
    }
    \label{tab:counting}

  \end{threeparttable}
\end{wraptable}

%% file: tables/ablate_mask.tex
\begin{table*}[t]
  \centering
  \begin{threeparttable}

    \resizebox{\linewidth}{!}{%
    \begin{tabular}{l SSSSSS S}
      \toprule
      \textbf{Method} & \textbf{RefCOCO} & \textbf{RefCOCO+} & \textbf{RefCOCOg} & \textbf{ReasonSeg} & \textbf{DRAM} & \textbf{Manga109} & \textbf{Average} \\
      \midrule
      Threshold (128) & 72.7 & 55.7 &  \textbf{76.7} & 72.8 & 77.5 &  \textbf{53.5} & 68.2 \\
      Threshold (225) & 72.5 & 55.5 & 76.4 & 72.8 & 77.5 &  \textbf{53.5} & 68.0 \\
      \rowcolor{gray!20} Otsu &  \textbf{73.1} &  \textbf{56.1} &  \textbf{76.7} &  \textbf{73.7} &  \textbf{79.6} & 53.3 &  \textbf{68.8} \\
      K-means         &  \textbf{73.1} & 52.8 &  \textbf{76.7} & 69.2 & 78.3 & 53.4 & 67.3 \\
      \bottomrule
    \end{tabular}
    }

    \vspace{5pt}
    \caption{
      \textbf{Ablation Study on Mask Extraction Methods for Segmentation Tasks}.
      We conduct an ablation study on Nano Banana 2 across six segmentation benchmarks (100-image subsets) and report the gIoU metric. The default setting is marked in \colorbox{gray!20}{grey}.
    }
    \label{tab:ablate_mask}
  \end{threeparttable}
\end{table*}

%% file: tables/ablate_prompt.tex
\begin{table}[t]
  \centering

  \resizebox{.85\linewidth}{!}{%
    \begin{tabular}{l c>{\columncolor{gray!20}}cc c c>{\columncolor{gray!20}}c c c>{\columncolor{gray!20}}c}
      \toprule
      \multirow{2}{*}{\textbf{Model}} & \multicolumn{3}{c}{\textbf{Seg. (RefCOCOg)}} && \multicolumn{2}{c}{\textbf{Depth (NYUv2)} $\downarrow$} && \multicolumn{2}{c}{\textbf{Counting (PixMo)} $\downarrow$}\\
      \cmidrule{2-4} \cmidrule{6-7} \cmidrule{9-10}
      & {short} & {mid} & {long} && {short} & {mid} && {short} & {mid} \\
      \midrule
      Nano Banana Pro & 74.8 & \textbf{75.2} & 74.8 && 11.5 & \textbf{11.0} && 5.1 & \textbf{4.7} \\
      Nano Banana 2   & \textbf{77.0} & 76.7 & 75.7 && 11.4 & \textbf{11.1} && 4.0 & \textbf{3.9} \\
      GPT-Image 2     & 24.6 & 52.5 & \textbf{57.9} && 10.9 & \textbf{9.6} && 5.9 & \textbf{5.5} \\
      Seedream-5.0    & 8.0  & \textbf{40.1} & 37.5 && {$\times$} & {$\times$} && {$\times$} & {$\times$} \\
      Qwen-Edit-2511  & 11.6 & \textbf{36.3} & 34.2 && 6.5 & \textbf{6.1} && {$\times$} & {$\times$} \\
      \bottomrule
    \end{tabular}
  }

  \vspace{5pt}
  \caption{
    \textbf{Ablation Study on Prompt Granularity},
    evaluated on 100-image subsets of RefCOCOg (gIoU), NYUv2 (AbsRel), and PixMo-Points (MAE).
    \colorbox{gray!20}{Grey shading} indicates the default configuration.
  }
  \label{tab:ablate_prompt}

\end{table}

%% file: figures/fig_bad_case1.tex
\begin{figure}[t]
  \centering
  \includegraphics[page=1, width=.95\linewidth]{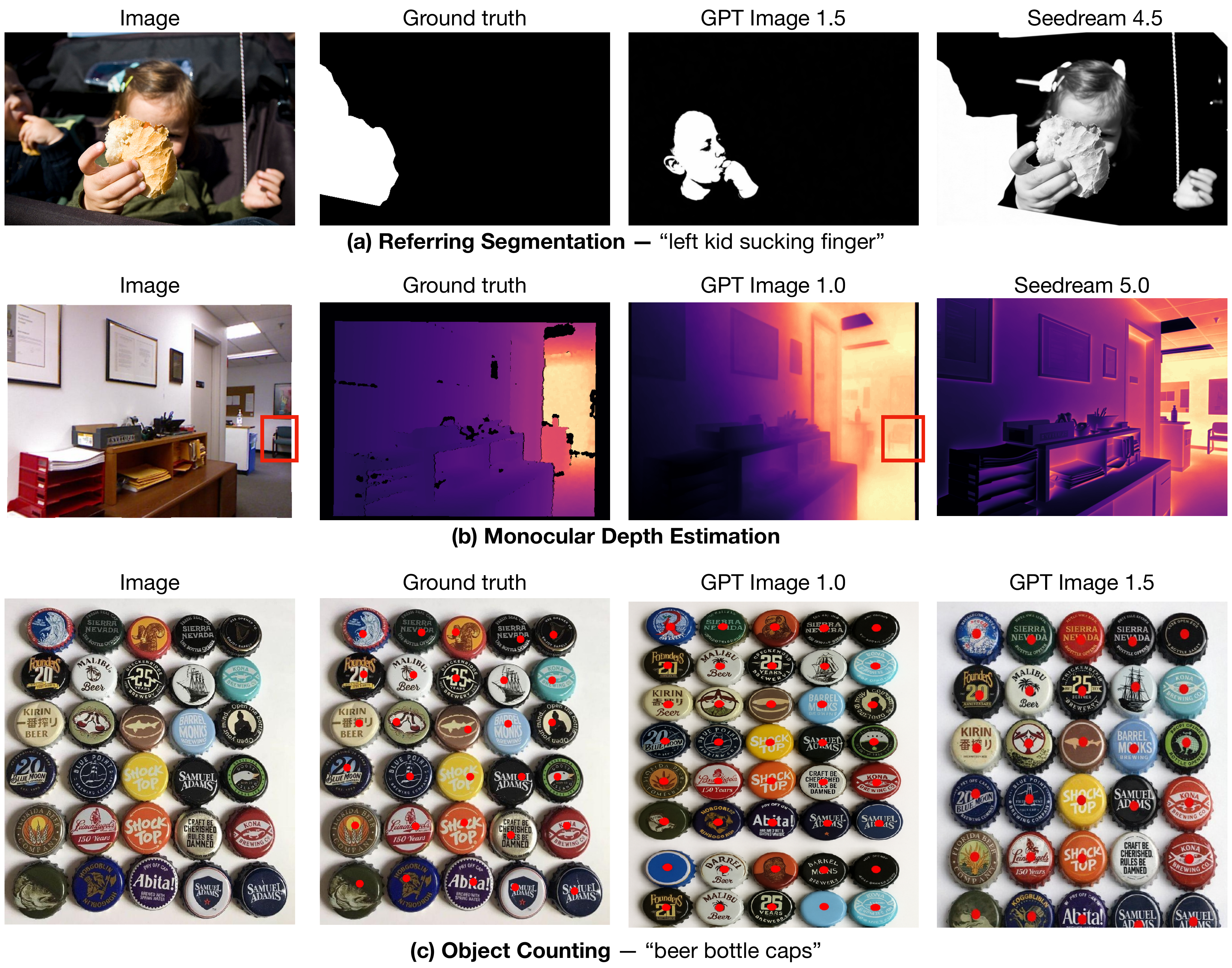}
  \vspace{-5pt}
  \caption{
    \textbf{Systematic Failure of GPT-Image-1.0/1.5 and Seedream Series.}
    GPT-Image-1.0/1.5 return outputs in the correct task format but fail to preserve the input, regenerating the scene instead of transforming it in place---from geometric drift that breaks pixel-wise correspondence (b) to wholesale re-rendering that hallucinates new content, such as the redrawn bottle-cap designs (c).
    The Seedream series, on the other hand, fails to follow the instruction, returning a stylized or near-unchanged copy of the input rather than performing the requested task (a,b).
  }\label{fig:bad_case1}
  \vspace{-5pt}
\end{figure}

%% file: figures/fig_bad_case3.tex
\begin{figure}[!t]
  \centering
  \includegraphics[width=.7\linewidth]{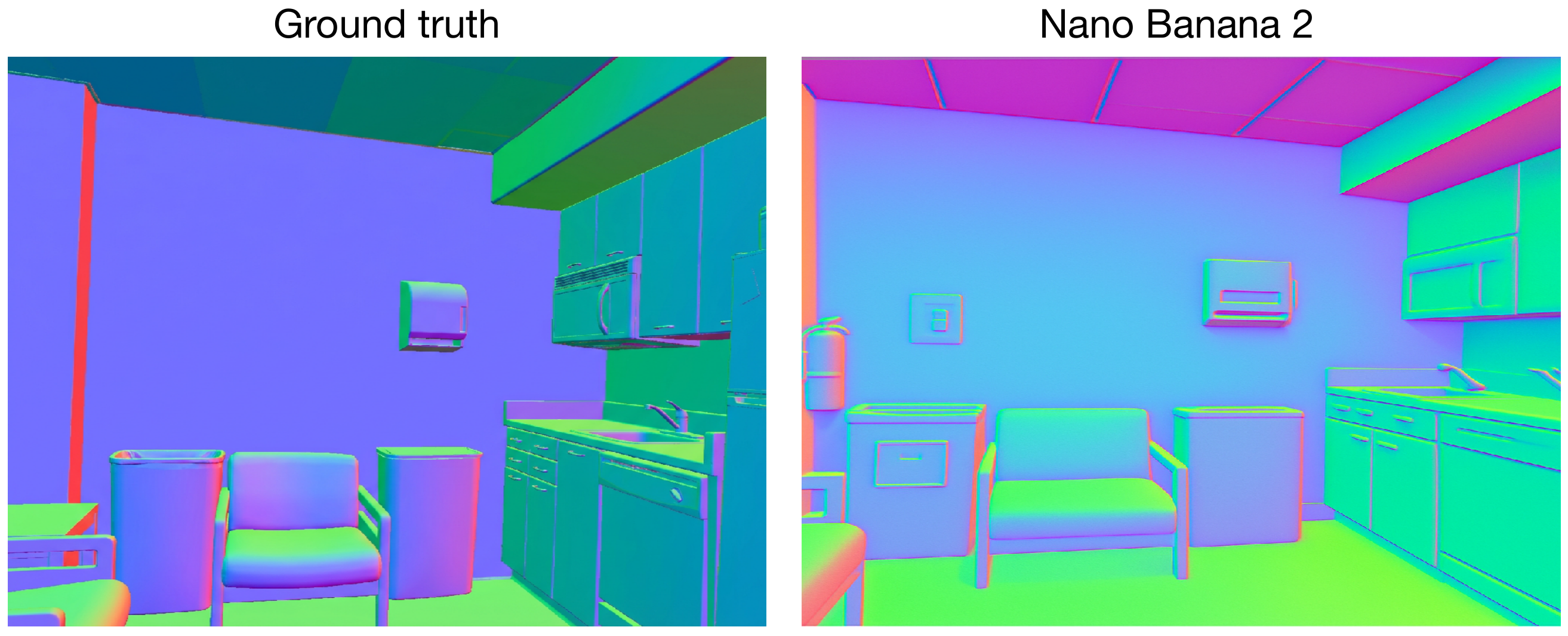}
  \vspace{-5pt}
  \caption{
    \textbf{Failure on Surface Normal Estimation.}
    Image generators produce normal-like outputs that may look plausible at a glance, but they do not preserve the correct scene geometry or faithfully follow the prescribed RGB encoding of surface normals.
  }\label{fig:bad_case3}
\end{figure}

%% file: sec/5_conclusion.tex
\section{Conclusion}
In this paper, we systematically evaluated general-purpose generative models as zero-shot visual perceivers. By re-purposing image synthesis for dense perception via text prompting, we showed that these models already encode non-trivial geometric and semantic understanding without task-specific fine-tuning.
Our benchmarking reveals a clear trade-off: specialized models remain stronger in in-distribution accuracy and efficiency, while zero-shot generative prompting is more competitive on OOD scenarios and reasoning-heavy tasks. Despite persistent challenges in alignment, output fidelity, and speed, these results point to a promising path towards unified architectures that bridge synthesis and perception. We will release \textsc{ProbeGen}, our evaluation suite and prompt templates, to support further research.


%% file: sec/X_supp.tex
\newpage
\appendix

\section*{Supplementary Material}

\noindent This supplementary material is organized as follows.
Appendix~\ref{sec:supp-counting} details the deterministic counting decoder $E_\text{count}$---how its decoding strategy was selected, the leakage-safe protocol that fixes its hyperparameters, and the residual error sources it cannot recover.
Appendix~\ref{sec:supp-efficiency} documents the efficiency measurement protocol and reports throughput and memory for all models, including the API-hosted ones excluded from the main tables.
Appendix~\ref{sec:supp-qualitative} provides additional qualitative comparisons across depth, segmentation, and counting, including out-of-distribution and reasoning examples, with an automated and logged example-selection criterion for each figure.
Appendix~\ref{sec:supp-limitations} discusses the current limitations of \textsc{ProbeGen} and the extensions they motivate.

\section{Counting Decoder Details}
\label{sec:supp-counting}
This supplementary expands on Sec.~\ref{sec:method-counting}, which introduced the deterministic decoder $E_\text{count}$ that maps a generator's marked image to an integer count. We describe (i) how we selected the decoding strategy, (ii) the leakage-safe protocol that sets its hyperparameters, and (iii) the residual error sources that the decoder cannot recover.

\subsection{Decoder Selection}

We compared eight candidate decoders that all reduce the same generated RGB image to a count but differ in how they isolate dots: connected-components on a red HSV mask (with various opening widths and an area-size window), a contour-based circularity filter, distance-transform peak detection, and the shape-filtered blob detector used in this paper, with the connected-components and blob decoders additionally evaluated under the diff-vs-input gate. Decoders were scored on the cached generations of three image generators (Nano Banana Pro, Nano Banana 2, and GPT-Image-2) so that the comparison is decode-only and re-running the (expensive) generators is unnecessary.

Table~\ref{tab:ablate_counting_decoder} reports MAE per generator. Two clear trends emerge. First, decoders without a shape model (connected-components on the red mask, with or without opening) systematically overcount: a single elongated red smear or a thin red contour is counted just like a true dot, and an MAE in the 30--60 range reflects this. Adding the diff-vs-input gate alone is insufficient: it sharply reduces overcounting for Nano Banana 2 (which edits largely in place, reaching 5.50 MAE) but leaves GPT-Image-2 at 10.90 MAE, since the latter re-renders the whole scene and weakens the diff signal. Second, the shape-filtered \texttt{\small SimpleBlobDetector} -- which requires each candidate to satisfy area, circularity, and inertia-ratio constraints -- dominates everywhere, and combining it with the diff gate (\texttt{\small diff\_red\_blob}) reduces the average MAE to 5.07. We therefore adopt \texttt{\small diff\_red\_blob} as the deployed decoder across all generators.

\input{tables/ablate_counting_decoder}

\subsection{Leakage-Safe Hyperparameter Sweep}

The blob filter has five hyperparameters: an HSV saturation floor $S$, minimum and maximum blob areas, minimum circularity, and minimum inertia ratio. To set them without overfitting, we grid-search on the pooled three-generator outputs (300 samples in total) using a split that is leakage-safe by construction.

\vspace{3pt} \noindent \textbf{Split protocol.}
The three model-versions of the same source image are highly correlated, since they share the underlying scene and instance layout. We therefore split by \textit{source image} rather than by instance: we shuffle the 100 image identifiers (seed 0), put 50 in train and 50 in val, and force all three generations of an image into the same fold. This prevents a scene from appearing in both train and val regardless of which generator produced it. Hyperparameters are selected on train MAE, and we report pooled val MAE.

\vspace{3pt} \noindent \textbf{Result.}
Table~\ref{tab:ablate_counting_hparam} summarizes the sweep. The literal argmin pushes \texttt{\small minArea}, \texttt{\small minInertia}, and \texttt{\small minCircularity} towards extreme values, which is the kind of choice we expect to overfit to small validation sets. We therefore adopt only the two robust, directionally-consistent improvements: raising the saturation floor from~150 to~160 (which more cleanly rejects washed-out reds) and the maximum blob area from~800 to~1500 (which keeps the filter forgiving of slightly oversized markers). The remaining shape floors are deliberately left at conservative values.

\input{tables/ablate_counting_hparam}

\subsection{Residual Failure Modes}

Two error sources are inherent to the mark-then-count probe and cannot be recovered by any decoder:

\vspace{3pt} \noindent \textbf{Red-cluttered scenes.}
When the queried category appears in a scene that already contains red structures ({\em e.g.}, a supermarket aisle of red packaging or a flower bed of red poppies), the generator paints into a background that the diff gate and the colour mask cannot reliably distinguish from a genuine marker. The shape filter suppresses most of this background but cannot eliminate it.

\vspace{3pt} \noindent \textbf{Merged adjacent dots.}
In densely packed scenes, two markers can land close enough that their painted regions touch and are recovered as a single blob, causing undercounting. A distance-transform-based decoder (\texttt{\small peak\_dt}) addresses this by splitting touching blobs, but it over-splits isolated dots and is a net loss on this distribution (Table~\ref{tab:ablate_counting_decoder}).

These limitations are generation-side rather than decoder-side: progress would require a different marking strategy ({\em e.g.}, alternating colours, or markers carried on a separate channel), not a better post-processing pipeline. The MAE values reported in the main paper should therefore be read as bounded from below by these residual generation artifacts.

\section{Efficiency Measurements}
\label{sec:supp-efficiency}

The main tables report throughput and GPU memory only for locally deployed models, since these are the only measurements obtained under a controlled protocol. Table~\ref{tab:supp-efficiency} consolidates the efficiency numbers of all evaluated models, including the API-hosted ones, and we describe how each was measured below.

\vspace{3pt} \noindent \textbf{Local models.}
All locally deployed models (specialists, Flux.2-dev, and Qwen-Edit-2511) are benchmarked on a single NVIDIA H800 GPU with batch size 1, using the same preprocessing as the main evaluation (inputs resized to a long side of 1024). After warm-up iterations, we time the full per-image inference (including any text-encoder or tool-calling steps, \eg the vLLM-served Qwen3-VL in the SAM3 pipelines) on a fixed subset of ReasonSeg for segmentation, NYUv2 for depth, and PixMo-Points for counting, and report the mean throughput in images per minute together with peak GPU memory.

\vspace{3pt} \noindent \textbf{API-hosted models.}
Proprietary generators and MLLMs are accessed exclusively through their APIs. For these we record the end-to-end wall-clock time of one request at a time (no batching or concurrency) and convert it to images per minute, marked with $^\dag$ in Table~\ref{tab:supp-efficiency}. This number includes network latency, server-side queueing, and any provider-side load balancing, and varies with the endpoint, region, and time of day; GPU memory is not observable. API response times are therefore reported for reference only and should not be compared with local throughput. Even so, they confirm the qualitative conclusion of the main paper: every image generator, hosted or local, delivers on the order of one to a few images per minute, one to three orders of magnitude below specialists.

\input{tables/supp_efficiency}

\section{Additional Qualitative Results}
\label{sec:supp-qualitative}

We provide additional qualitative comparisons across all three tasks, using a curated set of representative models alongside the ground truth. All predictions are post-processed exactly as in the main evaluation.
 To avoid cherry-picking, all examples are chosen by a logged, automated criterion (stated in each caption)---typically a mid-pack rather than a best case.

\vspace{3pt} \noindent \textbf{Depth and segmentation galleries.}
Figure~\ref{fig:supp-depth} compares zero-shot depth against discriminative and task-adapted specialists on NYUv2. Discriminative and task-adapted specialists (Depth Anything V2-L, Lotus-2) recover sharper, better-calibrated geometry, while the general generators produce coherent but softer relative depth without any depth-specific adaptation. 
Figure~\ref{fig:supp-refseg} shows referring and reasoning segmentation, where Nano Banana 2 grounds both direct references and multi-step reasoning queries zero-shot and remains competitive with the agentic SAM3\,$+$\,Qwen3-VL pipeline.

\vspace{3pt} \noindent \textbf{Out-of-distribution robustness.}
Figure~\ref{fig:supp-ood} highlights the regime where generators are most advantageous: On stylized domains that depart from natural-image statistics (Manga109 and DRAM), the general generators (Nano Banana Pro/2) preserve semantic binding and produce clean masks, whereas the specialists over- or under-segment. This qualitatively corroborates the OOD gap reported in Table~\ref{tab:reason_ood_seg}. 

\vspace{3pt} \noindent \textbf{Counting pipeline.}
Figure~\ref{fig:supp-counting} visualizes the mark-then-count interface end-to-end: the generator paints one red marker per instance and the deterministic decoder $E_\text{count}$ recovers the count, shown against the ground-truth points for both moderate and densely-packed scenes.

\input{figures/supp_gallery_depth}
\input{figures/supp_gallery_refseg}
\input{figures/supp_ood_seg}
\input{figures/supp_counting}

\section{Limitations and Future Work}
\label{sec:supp-limitations}

We view \textsc{ProbeGen} as a first step towards measuring zero-shot generative perception, and we outline its current limitations together with the natural extensions they motivate.

\vspace{3pt} \noindent \textbf{Evaluation scale and protocol.}
To keep API cost and wall-clock time tractable, we evaluate on a 500-image subset of each benchmark and use a single deterministic sample per image. This is sufficient to establish the qualitative orderings and trade-offs we report, but it widens confidence intervals and is one reason the Vision Banana numbers (computed on full benchmarks) are not directly comparable. Scaling to the full benchmarks and to multi-sample inference ({\em e.g.}, self-consistency over several generations) would tighten the estimates and is a natural extension of the protocol.

\vspace{3pt} \noindent \textbf{Generation-side bottlenecks.}
As discussed in the main paper, the dominant error sources are intrinsic to generation rather than to our extractors: imperfect pixel-wise alignment, output-format non-compliance, and the residual counting artefacts above (red-cluttered scenes and merged adjacent dots). Our deterministic extractors, applied identically across all models, are designed not to mask these effects, so the reported metrics are best read as lower bounds on what better-aligned generators could achieve.

\vspace{3pt} \noindent \textbf{Broader model and task coverage.}
\textsc{ProbeGen} currently spans three perception tasks and a fixed roster of generators. Beyond the models in our tables, we also tested four recent open-weight models through the same prompt-only protocol: the unified understanding-and-generation models BAGEL~\cite{bagel}, OmniGen2~\cite{omnigen2}, and Lance~\cite{lance}, and the in-context Flux variant UNO~\cite{uno}. Across the segmentation, depth, and counting prompts at all granularities (Figure~\ref{fig:prompt}), none of them produced a valid mask, depth map, or dot annotation: the outputs are the input image with a global, filter-like change (\eg a re-colouring or contrast shift), or a near-identical copy, and are indistinguishable from the instruction-following failures in Figure~\ref{fig:bad_case1}. Because no valid output exists to score, these models would receive $\times$ on every benchmark and we omit them from the tables (Sec.~\ref{sec:failure}). The lesson is that coupling understanding and generation natively does not by itself make perceptual outputs promptable; the missing ingredient is instruction-following for pixel-registered, structured outputs, a capability that lightweight instruction tuning (as in Vision Banana) is designed to instil, making these open-weight models natural targets for such tuning. The protocol extends directly to further open-weight generators (\eg EMU3.5, UniWorld, SenseNova-U1) and to additional structured tasks beyond depth, segmentation, and counting---surface normal estimation, which we found unsolved in zero-shot settings (Sec.~\ref{sec:failure}), is a particularly informative next target.

\vspace{3pt} \noindent \textbf{Beyond RGB output spaces.}
\textsc{ProbeGen} evaluates tasks whose outputs are naturally images (masks, depth maps) or can be rendered onto one (dots). Converting between structured annotations and RGB space is not free: the residual counting errors in Appendix~\ref{sec:supp-counting} arise exactly at this conversion, so any extension should keep the extractor $E_t$ deterministic and simple. With that constraint, tasks that are less naturally represented as images remain reachable through the same interface.
\emph{Bounding boxes} (object detection) require no new interface: the tight box of the Otsu-extracted mask already yields a detection, and, alternatively, the generator can be asked to draw a solid-colour rectangle that is recovered by colour isolation and a rectangularity test.
\emph{Keypoints} (human pose) reuse the mark-then-count interface with named, colour-coded markers---one colour per keypoint ({\em e.g.}, ``a red dot on the left wrist, a blue dot on the right wrist'')---decoded by running the per-colour blob detector of $E_\text{count}$ once per colour, so that the marker's colour identifies the keypoint and its centroid gives the location.
\emph{Text-valued outputs} (classification, attributes, visual question answering) can be obtained by asking the generator to write the answer onto the image and decoding it with OCR.
More broadly, our results show that the output interface, not just the model family, shapes accuracy ({\em e.g.}, MLLMs emit a scalar while generators must render it). Each task therefore admits alternative interfaces worth probing---counting via drawn boxes rather than dots, segmentation via a coloured overlay rather than a separate binary mask---and comparing interfaces would help disentangle a model's underlying perceptual competence from the difficulty of the chosen rendering format.

\vspace{3pt} \noindent \textbf{Richer inference settings.}
Finally, several models expose inference-time controls---such as explicit ``thinking'' or reasoning modes---whose effect on generative perception we do not yet ablate. Systematically varying these settings, alongside prompt and sampling strategies, is a low-cost way to extend the ablations in the main paper.

%% file: tables/ablate_counting_decoder.tex
\begin{table}[ht]
  \centering
  \begin{threeparttable}
    \resizebox{.85\linewidth}{!}{%
    \begin{tabular}{l l cccc}
      \toprule
      \multirow{2}{*}{\textbf{Decoder}} & \multirow{2}{*}{\textbf{Shape model}} & \multicolumn{4}{c}{\textbf{MAE} $\downarrow$} \\
      \cmidrule(lr){3-6}
      & & {NB Pro} & {NB 2} & {GPT-Image-2} & {Avg.} \\
      \midrule
      red\_open\_cc          & none (CC)                         & 34.73 & 37.99 & 41.32 & 38.01 \\
      red\_wide\_open\_cc    & none (CC)                     & 52.13 & 59.28 & 57.40 & 56.27 \\
      area\_window           & area only                                             & 42.33 & 47.59 & 50.55 & 46.82 \\
      peak\_dt               & distance   peaks                  & 67.73 & 73.16 & 76.74 & 72.54 \\
      circularity            & contour shape                     & 13.74 & 15.78 & 16.22 & 15.25 \\
      diff\_red\_open\_cc    & none (CC)   + diff                        &  9.34 &  5.50 & 10.90 &  8.58 \\
      blobdetector           & area + circ. + inertia                      &  5.62 &  5.36 &  6.28 &  5.75 \\
      \rowcolor{gray!20} diff\_red\_blob (ours) & area + circ. + inertia + diff & \textbf{5.22} & \textbf{4.55} & \textbf{5.43} & \textbf{5.07} \\
      \bottomrule
    \end{tabular}
    }
    \vspace{5pt}
    \caption{\textbf{Counting decoder ablation.} MAE on the 100-image PixMo-Points subset, evaluated on cached generations from three image generators (Nano Banana Pro, Nano Banana 2, GPT-Image-2); all decoders take the same RGB output and reduce it to a count, using fixed pre-sweep blob parameters.
    The default setting is highlighted in \colorbox{gray!20}{grey}.}
    \label{tab:ablate_counting_decoder}
  \end{threeparttable}
\end{table}

%% file: tables/ablate_counting_hparam.tex
\begin{table}[t]
  \centering
  \begin{threeparttable}
    \resizebox{.7\linewidth}{!}{
    \begin{tabular}{l c c l}
      \toprule
      \textbf{Hyperparameter} & \textbf{Before} & \textbf{After} & \textbf{Adopted?} \\
      \midrule
      Saturation floor $S$    & 150  & \textbf{160}    & Yes, robust improvement \\
      Max blob area           & 800  & \textbf{1500}   & Yes, robust improvement \\
      Min blob area           & 12   & 4 by argmin     & No, kept conservative \\
      Min circularity         & 0.6  & 0.7 by argmin   & No, kept conservative \\
      Min inertia ratio       & 0.5  & 0.2 by argmin   & No, kept conservative \\
      \midrule
      \multicolumn{4}{l}{\textit{Pooled validation MAE (50 held-out source images $\times$ 3 generators)}} \\
      \hspace{0.5em} Previous defaults         & \multicolumn{3}{l}{5.09} \\
      \hspace{0.5em} Literal argmin            & \multicolumn{3}{l}{4.61 (overfit risk)} \\
      \hspace{0.5em} \textbf{Adopted config}   & \multicolumn{3}{l}{\textbf{4.74}} \\
      \bottomrule
    \end{tabular}
    }
    \vspace{5pt}
    \caption{\textbf{Leakage-safe hyperparameter sweep for the blob filter.} The 100 source images are split 50/50 by image id (seed 0), forcing all three model-versions of an image into the same fold. Hyperparameters are tuned on train MAE; we report pooled val MAE.}
    \label{tab:ablate_counting_hparam}
  \end{threeparttable}
\end{table}

%% file: tables/supp_efficiency.tex
\begin{table}[t]
  \centering
  \begin{threeparttable}
    \resizebox{\linewidth}{!}{%
    \begin{tabular}{l l cc cc cc}
      \toprule
      \multirow{2}{*}{\textbf{Model}} & \multirow{2}{*}{\textbf{Deployment}} & \multicolumn{2}{c}{\textbf{Segmentation}} & \multicolumn{2}{c}{\textbf{Depth}} & \multicolumn{2}{c}{\textbf{Counting}} \\
      \cmidrule(lr){3-4} \cmidrule(lr){5-6} \cmidrule(lr){7-8}
      & & IMG/min & GB & IMG/min & GB & IMG/min & GB \\
      \midrule
      \textcolor{gray}{\textit{Specialists}} &&&&&&& \\
      \hspace{0.5em} SAM3 + Qwen3-VL-8B-Thinking~\cite{sam3}        & Local & 2.6 & 29  & {--} & {--} & {--} & {--} \\
      \hspace{0.5em} SAM3 + Qwen3-VL-32B-Thinking-FP8~\cite{sam3}   & Local & 2.1 & 73  & {--} & {--} & {--} & {--} \\
      \hspace{0.5em} Qwen3-VL-8B-SAMTok~\cite{samtok}               & Local & 54  & 19  & {--} & {--} & {--} & {--} \\
      \hspace{0.5em} Depth Anything V2-Large~\cite{da2}             & Local & {--} & {--} & 828 & 3.5 & {--} & {--} \\
      \hspace{0.5em} CountGD++~\cite{countgd++}                     & Local & {--} & {--} & {--} & {--} & 471 & 4.1 \\
      \hspace{0.5em} Marigold v1.1~\cite{marigold}                  & Local & {--} & {--} & 226 & 10  & {--} & {--} \\
      \hspace{0.5em} Lotus-2~\cite{lotus2}                          & Local & {--} & {--} & 25  & 34  & {--} & {--} \\
      \midrule
      \textcolor{gray}{\textit{MLLMs}} &&&&&&& \\
      \hspace{0.5em} Gemini-3.5-Flash~\cite{gemini_3_5}             & API   & {--} & {--} & {--} & {--} & 9.2$^\dag$  & {--} \\
      \hspace{0.5em} GPT-5.4~\cite{gpt_5_4}                         & API   & {--} & {--} & {--} & {--} & 22.6$^\dag$ & {--} \\
      \hspace{0.5em} Claude-Opus-4.7~\cite{claude_opus_4_7}         & API   & {--} & {--} & {--} & {--} & 13.0$^\dag$ & {--} \\
      \midrule
      \textcolor{gray}{\textit{Generative Generalists}} &&&&&&& \\
      \hspace{0.5em} Nano Banana Pro~\cite{nano_banana_pro}         & API   & 1.2$^\dag$ & {--} & 1.2$^\dag$ & {--} & 2.8$^\dag$ & {--} \\
      \hspace{0.5em} Nano Banana 2~\cite{nano_banana_2}             & API   & 2.0$^\dag$ & {--} & 1.9$^\dag$ & {--} & 2.4$^\dag$ & {--} \\
      \hspace{0.5em} GPT-Image-2~\cite{gpt_image_2}                 & API   & 1.5$^\dag$ & {--} & 1.5$^\dag$ & {--} & 1.1$^\dag$ & {--} \\
      \hspace{0.5em} Seedream-4.5~\cite{seedream_4_5}               & API   & {$\times$} & {--} & 3.3$^\dag$ & {--} & {$\times$} & {--} \\
      \hspace{0.5em} Seedream-5.0~\cite{seedream_5_0}               & API   & 1.6$^\dag$ & {--} & {$\times$} & {--} & {$\times$} & {--} \\
      \hspace{0.5em} Flux.2-dev-NF4 (32B)~\cite{flux-2}             & Local & 1.2 & 39  & {$\times$} & {$\times$} & {$\times$} & {$\times$} \\
      \hspace{0.5em} Qwen-Edit-2511 (20B)~\cite{qwen-image}         & Local & 1.5 & 61  & 1.5 & 62 & {$\times$} & {$\times$} \\
      \bottomrule
    \end{tabular}
    }
    \vspace{5pt}
    \caption{
      \textbf{Efficiency of all evaluated models.}
      Throughput (IMG/min) and peak GPU memory (GB) of locally deployed models, measured on a single NVIDIA H800 (Appendix~\ref{sec:supp-efficiency}).
      $^\dag$: end-to-end API response time converted to IMG/min, which includes network latency and server-side queueing and is therefore \emph{not} comparable to local throughput; GPU memory of API-hosted models is unavailable.
      $\times$: no valid output for the task; {--}: not applicable or unavailable.
    }
    \label{tab:supp-efficiency}
  \end{threeparttable}
\end{table}

%% file: figures/supp_gallery_depth.tex
\begin{figure}[t]
  \centering
  \includegraphics[width=\linewidth]{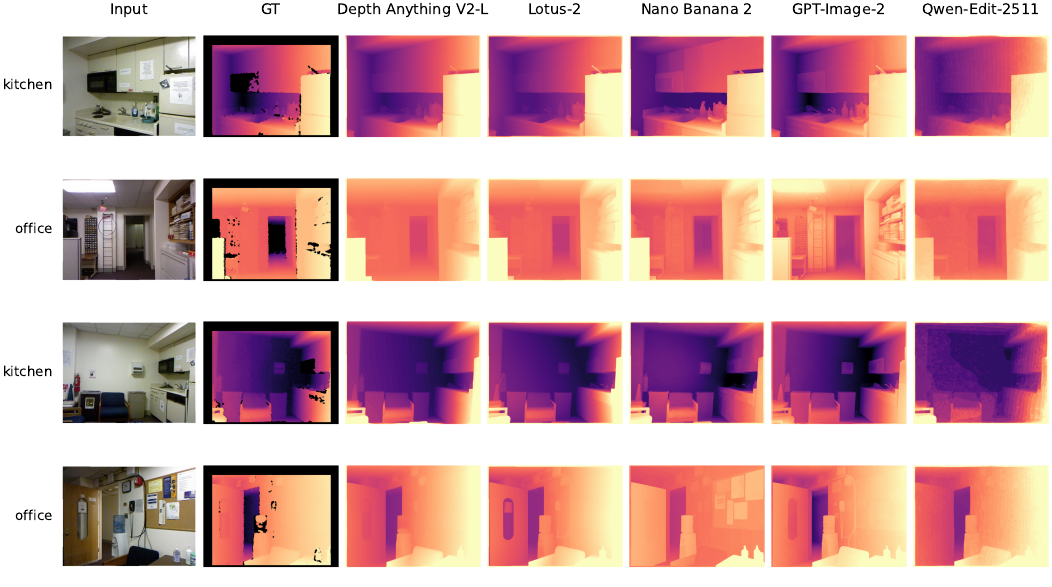}
  \vspace{-5mm}
  \caption{
    \textbf{Zero-shot depth qualitative comparison on NYUv2.}
    Predicted depth is affine-aligned to the ground truth and rendered with a shared
    per-row \texttt{magma} colormap (brighter\,$=$\,closer). 
    Scenes are selected as mid-pack (median-AbsRel) examples per
    room type, not best cases.
  }
  \label{fig:supp-depth}
\end{figure}

%% file: figures/supp_gallery_refseg.tex
\begin{figure}[t]
  \centering
  \includegraphics[width=\linewidth]{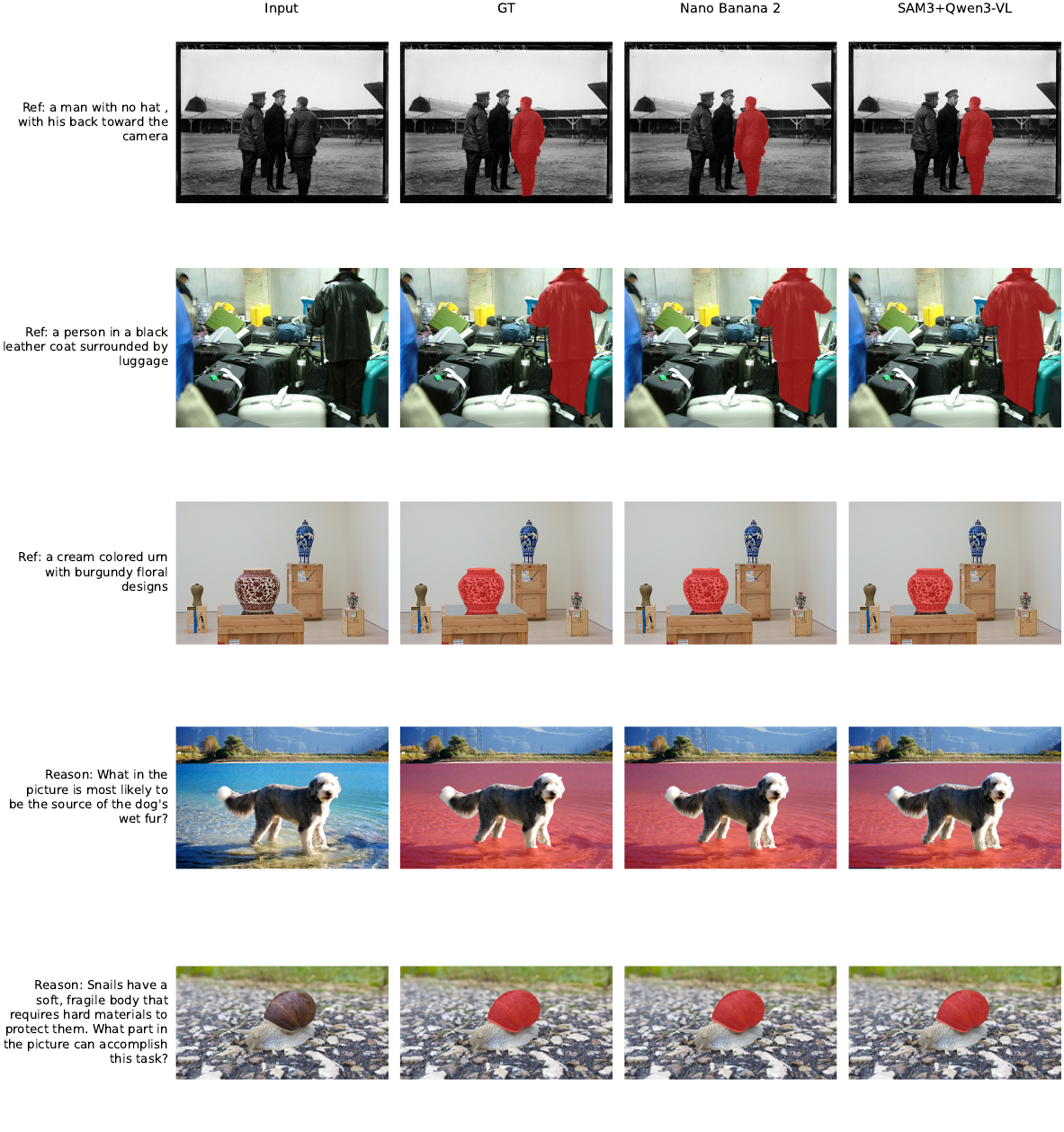}
  \caption{
    \textbf{Referring and reasoning segmentation qualitative comparison.}
    Top rows: RefCOCOg referring expressions; bottom rows: ReasonSeg reasoning queries
    ({\em e.g.}, ``the source of the dairy we eat'' $\rightarrow$ the cow). Masks are
    overlaid on the input for visualization (evaluation uses the binary masks).
    The RefCOCOg rows are high-IoU referring expressions; the ReasonSeg rows are high-IoU among multi-word reasoning queries
    (both ranked by Nano Banana 2 IoU, excluding the very top ranks).
  }
  \label{fig:supp-refseg}
\end{figure}

%% file: figures/supp_ood_seg.tex
\begin{figure}[t]
  \centering
  \includegraphics[width=\linewidth]{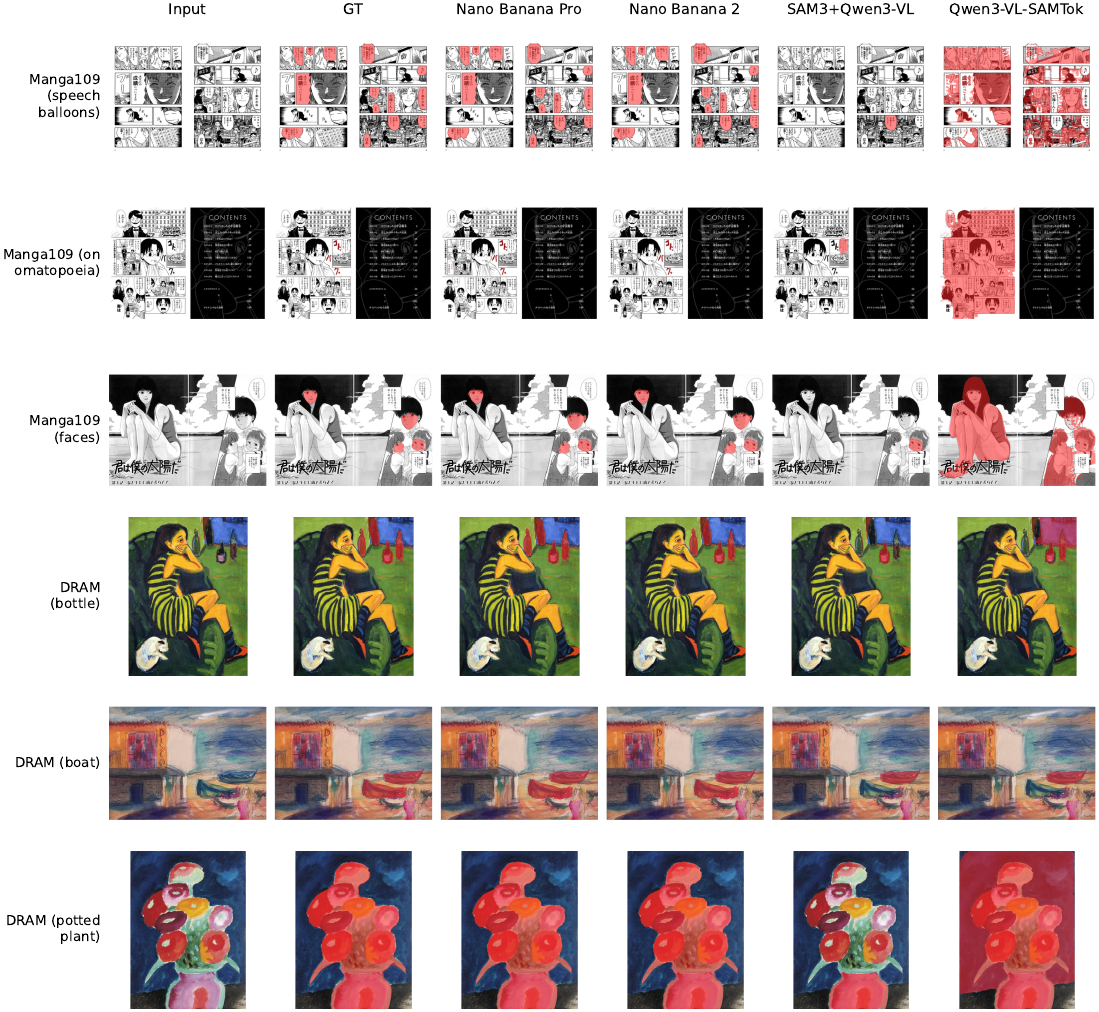}
  \caption{
    \textbf{Out-of-distribution segmentation showcase (Manga109 and DRAM).}
    Masks are recoloured on the input for visualization; evaluation uses the
    binary masks. 
    Rows are above-median generator-advantage cases (generators beat both
    specialists by IoU), sampled across object classes rather than top-1 per class.
  }
  \label{fig:supp-ood}
\end{figure}

%% file: figures/supp_counting.tex
\begin{figure}[t]
  \centering
  \includegraphics[width=\linewidth]{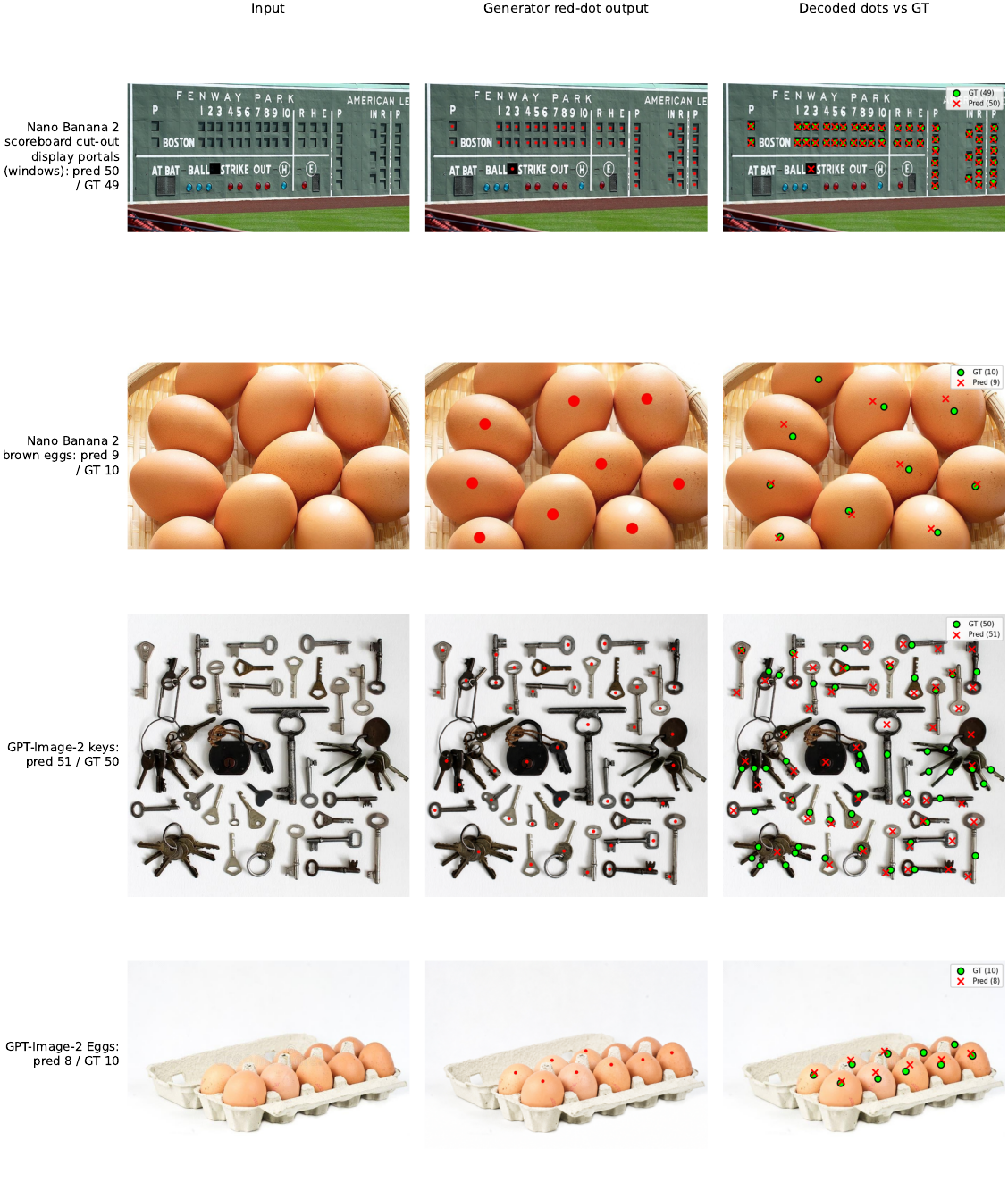}
  \caption{
    \textbf{The mark-then-count counting pipeline.}
    For each example we show the input, the generator's red-dot output, and the dots
    recovered by the deterministic decoder ($E_\text{count}$, red $\times$) overlaid against
    the ground-truth points (lime $\circ$). The generators place one marker per instance and
    the decoder recovers the count accurately on both moderate and densely-packed scenes.
    Examples are accurate-decode cases ($|\hat{c}-c|\!\le\!2$), including a crowded scene per
    model to show robustness.
  }
  \label{fig:supp-counting}
\end{figure}

%% file: bmvc_review.bbl
\begin{thebibliography}{43}
\providecommand{\natexlab}[1]{#1}
\providecommand{\url}[1]{\texttt{#1}}
\expandafter\ifx\csname urlstyle\endcsname\relax
  \providecommand{\doi}[1]{doi: #1}\else
  \providecommand{\doi}{doi: \begingroup \urlstyle{rm}\Url}\fi

\bibitem[Amini-Naieni and Zisserman(2026)]{countgd++}
Niki Amini-Naieni and Andrew Zisserman.
\newblock Countgd++: Generalized prompting for open-world counting.
\newblock In \emph{CVPR}, 2026.

\bibitem[Amini-Naieni et~al.(2024)Amini-Naieni, Han, and Zisserman]{countgd}
Niki Amini-Naieni, Tengda Han, and Andrew Zisserman.
\newblock Countgd: Multi-modal open-world counting.
\newblock In \emph{NeurIPS}, 2024.

\bibitem[{Anthropic}(2026)]{claude_opus_4_7}
{Anthropic}.
\newblock Claude opus 4.7.
\newblock \url{https://www.anthropic.com/news/claude-opus-4-7}, 2026.

\bibitem[Bai et~al.(2025)Bai, Cai, Chen, Chen, Chen, Cheng, Deng, Ding, Gao,
  Ge, et~al.]{qwen3-vl}
Shuai Bai, Yuxuan Cai, Ruizhe Chen, Keqin Chen, Xionghui Chen, Zesen Cheng,
  Lianghao Deng, Wei Ding, Chang Gao, Chunjiang Ge, et~al.
\newblock Qwen3-vl technical report.
\newblock \emph{arXiv preprint arXiv:2511.21631}, 2025.

\bibitem[Carion et~al.(2025)Carion, Gustafson, Hu, Debnath, Hu, Suris, Ryali,
  Alwala, Khedr, Huang, et~al.]{sam3}
Nicolas Carion, Laura Gustafson, Yuan-Ting Hu, Shoubhik Debnath, Ronghang Hu,
  Didac Suris, Chaitanya Ryali, Kalyan~Vasudev Alwala, Haitham Khedr, Andrew
  Huang, et~al.
\newblock Sam 3: Segment anything with concepts.
\newblock \emph{arXiv preprint arXiv:2511.16719}, 2025.

\bibitem[Cohen et~al.(2022)Cohen, Newman, and Shamir]{DRAM}
Nadav Cohen, Yael Newman, and Ariel Shamir.
\newblock {Semantic Segmentation in Art Paintings}.
\newblock \emph{Computer Graphics Forum}, 2022.

\bibitem[Dai et~al.(2017)Dai, Chang, Savva, Halber, Funkhouser, and
  Nie{\ss}ner]{scannet}
Angela Dai, Angel~X Chang, Manolis Savva, Maciej Halber, Thomas Funkhouser, and
  Matthias Nie{\ss}ner.
\newblock Scannet: Richly-annotated 3d reconstructions of indoor scenes.
\newblock In \emph{CVPR}, 2017.

\bibitem[{Deepmind}(2026)]{gemini_3_5}
{Deepmind}.
\newblock Gemini 3.5 flash.
\newblock \url{https://deepmind.google/models/gemini/flash/}, 2026.

\bibitem[Deitke et~al.(2025)Deitke, Clark, Lee, Tripathi, Yang, Park, Salehi,
  Muennighoff, Lo, Soldaini, et~al.]{pixmo}
Matt Deitke, Christopher Clark, Sangho Lee, Rohun Tripathi, Yue Yang, Jae~Sung
  Park, Mohammadreza Salehi, Niklas Muennighoff, Kyle Lo, Luca Soldaini, et~al.
\newblock Molmo and pixmo: Open weights and open data for state-of-the-art
  vision-language models.
\newblock In \emph{CVPR}, 2025.

\bibitem[Deng et~al.(2025)Deng, Zhu, Li, Gou, Li, Wang, Zhong, Yu, Nie, Song,
  Shi, and Fan]{bagel}
Chaorui Deng, Deyao Zhu, Kunchang Li, Chenhui Gou, Feng Li, Zeyu Wang, Shu
  Zhong, Weihao Yu, Xiaonan Nie, Ziang Song, Guang Shi, and Haoqi Fan.
\newblock Emerging properties in unified multimodal pretraining.
\newblock \emph{arXiv preprint arXiv:2505.14683}, 2025.

\bibitem[Fu et~al.(2026)Fu, Huang, Wu, Jiang, Huo, Li, Song, Ding, Guo, He, Fu,
  Mao, and Zhang]{lance}
Fengyi Fu, Mengqi Huang, Shaojin Wu, Yunsheng Jiang, Yufei Huo, Hao Li,
  Yinghang Song, Fei Ding, Jianzhu Guo, Qian He, Zheren Fu, Zhendong Mao, and
  Yongdong Zhang.
\newblock Lance: Unified multimodal modeling by multi-task synergy.
\newblock \emph{arXiv preprint arXiv:2605.18678}, 2026.

\bibitem[Gabeur et~al.(2026)Gabeur, Long, Peng, Voigtlaender, Sun, Bao, Truong,
  Wang, Zhou, Barron, et~al.]{image_generators_generalist_vision}
Valentin Gabeur, Shangbang Long, Songyou Peng, Paul Voigtlaender, Shuyang Sun,
  Yanan Bao, Karen Truong, Zhicheng Wang, Wenlei Zhou, Jonathan~T Barron,
  et~al.
\newblock Image generators are generalist vision learners.
\newblock \emph{arXiv preprint arXiv:2604.20329}, 2026.

\bibitem[Geiger et~al.(2012)Geiger, Lenz, and Urtasun]{kitti}
Andreas Geiger, Philip Lenz, and Raquel Urtasun.
\newblock Are we ready for autonomous driving? the kitti vision benchmark
  suite.
\newblock In \emph{CVPR}, 2012.

\bibitem[{Google DeepMind}(2025)]{nano_banana_pro}
{Google DeepMind}.
\newblock Nano banana pro.
\newblock \url{https://deepmind.google/models/gemini-image/pro/}, 2025.

\bibitem[{Google DeepMind}(2026)]{nano_banana_2}
{Google DeepMind}.
\newblock Nano banana 2.
\newblock \url{https://deepmind.google/models/gemini-image/flash/}, 2026.

\bibitem[He et~al.(2025)He, Li, Sheng, and Chen]{lotus2}
Jing He, Haodong Li, Mingzhi Sheng, and Ying-Cong Chen.
\newblock Lotus-2: Advancing geometric dense prediction with powerful image
  generative model.
\newblock \emph{arXiv preprint arXiv:2512.01030}, 2025.

\bibitem[Kazemzadeh et~al.(2014)Kazemzadeh, Ordonez, Matten, and Berg]{refcoco}
Sahar Kazemzadeh, Vicente Ordonez, Mark Matten, and Tamara Berg.
\newblock Referitgame: Referring to objects in photographs of natural scenes.
\newblock In \emph{EMNLP}, 2014.

\bibitem[Ke et~al.(2025)Ke, Qu, Wang, Metzger, Huang, Li, Obukhov, and
  Schindler]{marigold}
Bingxin Ke, Kevin Qu, Tianfu Wang, Nando Metzger, Shengyu Huang, Bo~Li, Anton
  Obukhov, and Konrad Schindler.
\newblock Marigold: Affordable adaptation of diffusion-based image generators
  for image analysis.
\newblock \emph{IEEE TPAMI}, 2025.

\bibitem[Kirillov et~al.(2023)Kirillov, Mintun, Ravi, Mao, Rolland, Gustafson,
  Xiao, Whitehead, Berg, Lo, et~al.]{sam}
Alexander Kirillov, Eric Mintun, Nikhila Ravi, Hanzi Mao, Chloe Rolland, Laura
  Gustafson, Tete Xiao, Spencer Whitehead, Alexander~C Berg, Wan-Yen Lo, et~al.
\newblock Segment anything.
\newblock In \emph{ICCV}, 2023.

\bibitem[Labs(2025)]{flux-2}
Black~Forest Labs.
\newblock {FLUX.2: Frontier Visual Intelligence}.
\newblock \url{https://bfl.ai/blog/flux-2}, 2025.

\bibitem[Lai et~al.(2024)Lai, Tian, Chen, Li, Yuan, Liu, and Jia]{lisa}
Xin Lai, Zhuotao Tian, Yukang Chen, Yanwei Li, Yuhui Yuan, Shu Liu, and Jiaya
  Jia.
\newblock Lisa: Reasoning segmentation via large language model.
\newblock In \emph{CVPR}, 2024.

\bibitem[Lin et~al.(2026)Lin, Chen, Liew, Chen, Li, Shi, Feng, and Kang]{da3}
Haotong Lin, Sili Chen, Junhao Liew, Donny~Y Chen, Zhenyu Li, Guang Shi, Jiashi
  Feng, and Bingyi Kang.
\newblock Depth anything 3: Recovering the visual space from any views.
\newblock In \emph{ICLR}, 2026.

\bibitem[Liu et~al.(2026)Liu, Wu, and Xie]{liu2026count}
Chang Liu, Haoning Wu, and Weidi Xie.
\newblock Count anything at any granularity.
\newblock \emph{arXiv preprint arXiv:2605.10887}, 2026.

\bibitem[OpenAI(2025)]{GPT-Image-1}
OpenAI.
\newblock Gpt-image-1, 2025.

\bibitem[{OpenAI}(2025)]{gpt_image_1_5}
{OpenAI}.
\newblock Gpt-image-1.5.
\newblock \url{https://openai.com/index/new-chatgpt-images-is-here/}, 2025.

\bibitem[{OpenAI}(2026{\natexlab{a}})]{gpt_5_4}
{OpenAI}.
\newblock Gpt 5.4.
\newblock \url{https://openai.com/index/introducing-gpt-5-4/},
  2026{\natexlab{a}}.

\bibitem[{OpenAI}(2026{\natexlab{b}})]{gpt_image_2}
{OpenAI}.
\newblock Gpt-image-2.
\newblock \url{https://openai.com/index/introducing-chatgpt-images-2-0/},
  2026{\natexlab{b}}.

\bibitem[Otsu et~al.(1979)]{otsu1979threshold}
Nobuyuki Otsu et~al.
\newblock A threshold selection method from gray-level histograms.
\newblock \emph{Automatica}, 11\penalty0 (285-296), 1979.

\bibitem[Ranftl et~al.(2020)Ranftl, Lasinger, Hafner, Schindler, and
  Koltun]{midas}
Ren{\'e} Ranftl, Katrin Lasinger, David Hafner, Konrad Schindler, and Vladlen
  Koltun.
\newblock Towards robust monocular depth estimation: Mixing datasets for
  zero-shot cross-dataset transfer.
\newblock \emph{IEEE TPAMI}, 2020.

\bibitem[Ravi et~al.(2025)Ravi, Gabeur, Hu, Hu, Ryali, Ma, Khedr, R{\"a}dle,
  Rolland, Gustafson, et~al.]{sam2}
Nikhila Ravi, Valentin Gabeur, Yuan-Ting Hu, Ronghang Hu, Chaitanya Ryali,
  Tengyu Ma, Haitham Khedr, Roman R{\"a}dle, Chloe Rolland, Laura Gustafson,
  et~al.
\newblock Sam 2: Segment anything in images and videos.
\newblock In \emph{ICLR}, 2025.

\bibitem[{Seedream Team}(2025)]{seedream_4_5}
{Seedream Team}.
\newblock Seedream 4.5.
\newblock \url{https://seed.bytedance.com/en/seedream4_5}, 2025.

\bibitem[{Seedream Team}(2026)]{seedream_5_0}
{Seedream Team}.
\newblock Seedream 5.0 lite.
\newblock \url{https://seed.bytedance.com/en/seedream5_0_lite}, 2026.

\bibitem[Silberman et~al.(2012)Silberman, Hoiem, Kohli, and Fergus]{nyuv2}
Nathan Silberman, Derek Hoiem, Pushmeet Kohli, and Rob Fergus.
\newblock Indoor segmentation and support inference from rgbd images.
\newblock In \emph{ECCV}, 2012.

\bibitem[Vasiljevic et~al.(2019)Vasiljevic, Kolkin, Zhang, Luo, Wang, Dai,
  Daniele, Mostajabi, Basart, Walter, et~al.]{diode}
Igor Vasiljevic, Nick Kolkin, Shanyi Zhang, Ruotian Luo, Haochen Wang, Falcon~Z
  Dai, Andrea~F Daniele, Mohammadreza Mostajabi, Steven Basart, Matthew~R
  Walter, et~al.
\newblock Diode: A dense indoor and outdoor depth dataset.
\newblock \emph{arXiv preprint arXiv:1908.00463}, 2019.

\bibitem[Wiedemer et~al.(2025)Wiedemer, Li, Vicol, Gu, Matarese, Swersky, Kim,
  Jaini, and Geirhos]{video_zero_shot_reasoners}
Thadd{\"a}us Wiedemer, Yuxuan Li, Paul Vicol, Shixiang~Shane Gu, Nick Matarese,
  Kevin Swersky, Been Kim, Priyank Jaini, and Robert Geirhos.
\newblock Video models are zero-shot learners and reasoners.
\newblock \emph{arXiv preprint arXiv:2509.20328}, 2025.

\bibitem[Wu et~al.(2025{\natexlab{a}})Wu, Li, Zhou, Lin, Gao, Yan, ming Yin,
  Bai, Xu, Chen, Chen, Tang, Zhang, Wang, Yang, Yu, Cheng, Liu, Li, Zhang,
  Meng, Wei, Ni, Chen, Cao, Peng, Qu, Wu, Wang, Yu, Wen, Feng, Xu, Wang, Zhang,
  Zhu, Wu, Cai, and Liu]{qwen-image}
Chenfei Wu, Jiahao Li, Jingren Zhou, Junyang Lin, Kaiyuan Gao, Kun Yan, Sheng
  ming Yin, Shuai Bai, Xiao Xu, Yilei Chen, Yuxiang Chen, Zecheng Tang, Zekai
  Zhang, Zhengyi Wang, An~Yang, Bowen Yu, Chen Cheng, Dayiheng Liu, Deqing Li,
  Hang Zhang, Hao Meng, Hu~Wei, Jingyuan Ni, Kai Chen, Kuan Cao, Liang Peng,
  Lin Qu, Minggang Wu, Peng Wang, Shuting Yu, Tingkun Wen, Wensen Feng,
  Xiaoxiao Xu, Yi~Wang, Yichang Zhang, Yongqiang Zhu, Yujia Wu, Yuxuan Cai, and
  Zenan Liu.
\newblock Qwen-image technical report.
\newblock \emph{arXiv preprint arXiv:2508.02324}, 2025{\natexlab{a}}.

\bibitem[Wu et~al.(2025{\natexlab{b}})Wu, Zheng, Yan, Xiao, Luo, Wang, Li,
  Jiang, Liu, Zhou, Liu, Xia, Li, Deng, Wang, Luo, Zhang, Lian, Wang, Wang,
  Huang, and Liu]{omnigen2}
Chenyuan Wu, Pengfei Zheng, Ruiran Yan, Shitao Xiao, Xin Luo, Yueze Wang, Wanli
  Li, Xiyan Jiang, Yexin Liu, Junjie Zhou, Ze~Liu, Ziyi Xia, Chaofan Li, Haoge
  Deng, Jiahao Wang, Kun Luo, Bo~Zhang, Defu Lian, Xinlong Wang, Zhongyuan
  Wang, Tiejun Huang, and Zheng Liu.
\newblock Omnigen2: Exploration to advanced multimodal generation.
\newblock \emph{arXiv preprint arXiv:2506.18871}, 2025{\natexlab{b}}.

\bibitem[Wu et~al.(2025{\natexlab{c}})Wu, Huang, Wu, Cheng, Ding, and He]{uno}
Shaojin Wu, Mengqi Huang, Wenxu Wu, Yufeng Cheng, Fei Ding, and Qian He.
\newblock Less-to-more generalization: Unlocking more controllability by
  in-context generation.
\newblock In \emph{ICCV}, 2025{\natexlab{c}}.

\bibitem[Xie et~al.(2025)Xie, Lin, Liu, Li, and Wong]{manga}
Minshan Xie, Jian Lin, Hanyuan Liu, Chengze Li, and Tien-Tsin Wong.
\newblock Advancing manga analysis: Comprehensive segmentation annotations for
  the manga109 dataset.
\newblock In \emph{CVPR}, 2025.

\bibitem[Yang et~al.(2024{\natexlab{a}})Yang, Kang, Huang, Xu, Feng, and
  Zhao]{da}
Lihe Yang, Bingyi Kang, Zilong Huang, Xiaogang Xu, Jiashi Feng, and Hengshuang
  Zhao.
\newblock Depth anything: Unleashing the power of large-scale unlabeled data.
\newblock In \emph{CVPR}, 2024{\natexlab{a}}.

\bibitem[Yang et~al.(2024{\natexlab{b}})Yang, Kang, Huang, Zhao, Xu, Feng, and
  Zhao]{da2}
Lihe Yang, Bingyi Kang, Zilong Huang, Zhen Zhao, Xiaogang Xu, Jiashi Feng, and
  Hengshuang Zhao.
\newblock Depth anything v2.
\newblock In \emph{NeurIPS}, 2024{\natexlab{b}}.

\bibitem[Yu et~al.(2016)Yu, Poirson, Yang, Berg, and Berg]{refcocog}
Licheng Yu, Patrick Poirson, Shan Yang, Alexander~C Berg, and Tamara~L Berg.
\newblock Modeling context in referring expressions.
\newblock In \emph{ECCV}, 2016.

\bibitem[Zhou et~al.(2026)Zhou, Zhang, Gong, Wu, Tian, Wang, Yuan, Wang, Qi,
  Fei, et~al.]{samtok}
Yikang Zhou, Tao Zhang, Dengxian Gong, Yuanzheng Wu, Ye~Tian, Haochen Wang,
  Haobo Yuan, Jiacong Wang, Lu~Qi, Hao Fei, et~al.
\newblock Samtok: Representing any mask with two words.
\newblock \emph{arXiv preprint arXiv:2601.16093}, 2026.

\end{thebibliography}
